\documentclass[10pt,twocolumn,letterpaper]{article}

\usepackage[pagenumbers]{cvpr}      % To produce the REVIEW version
\usepackage[breakable]{tcolorbox} % <-- Add [breakable] here
\usepackage{xcolor}
\definecolor{lightblue}{RGB}{115, 192, 222}
\definecolor{lightorange}{RGB}{253, 208, 162}

\newtcolorbox{promptbox}[3][Judge Prompt]{
  colback=black!5!white, % Light gray background color
  arc=5pt,               % Rounded corners
  boxrule=0.5pt,           % Border thickness
  fonttitle=\bfseries,     % Bold title
  title=#1,              % The title, uses the first optional argument
  colframe=#2,           % The frame color, uses the second argument
  label=#3,              % The label for cross-referencing, uses the third argument
  before upper={\small}, % Make the text inside a bit smaller
  fontupper=\fontfamily{ptm}\selectfont, % Sets the font (optional)
}
\newtcolorbox{promptbox*}[3][Judge Prompt]{
  breakable,
  float*=ht,
  width=\textwidth,
  colback=black!5,
  colframe=#2,
  arc=5pt,
  boxrule=0.5pt,
  fonttitle=\bfseries,
  title=#1,
  before upper={\small},
  fontupper=\fontfamily{ptm}\selectfont,
  label=#3,
}

\definecolor{cvprblue}{rgb}{0.21,0.49,0.74}
\usepackage[pagebackref,breaklinks,colorlinks,allcolors=cvprblue]{hyperref}
\usepackage{booktabs}
\usepackage{multirow}
\usepackage{array}
\usepackage{fancyvrb}
\usepackage{listings}
\usepackage{diagbox}
\usepackage{makecell}
\usepackage[table,xcdraw]{xcolor}
\usepackage{colortbl}  
\usepackage{fontawesome5}

\usepackage{tcolorbox} % Required for creating colored boxes
\usepackage{xcolor}    % Required for defining custom colors
\def\paperID{1978} % *** Enter the Paper ID here
\def\confName{CVPR}
\def\confYear{2026}

\title{FingerCap: Fine-grained Finger-level Hand Motion Captioning}

\author{
Xin Shen\textsuperscript{1,2} \quad
Rui Zhu\textsuperscript{1,3} \quad
Lei Shen\textsuperscript{4} \quad
Xinyu Wang\textsuperscript{2} \quad
Kaihao Zhang\textsuperscript{6} \quad
Tianqiang Zhu\textsuperscript{7} \\[0.3em]
Shuchen Wu\textsuperscript{1} \quad
Chenxi Miao\textsuperscript{1} \quad
Weikang Li\textsuperscript{5} \quad
Yang Li\textsuperscript{1}\thanks{Corresponding author.} \quad
Deguo Xia\textsuperscript{1} \quad
Jizhou Huang\textsuperscript{1} \quad
Xin Yu\textsuperscript{2}
\\[0.7em]
\textsuperscript{1}Baidu Inc. \\
\textsuperscript{2}The University of Queensland \quad
\textsuperscript{3}Nanjing University \quad
\textsuperscript{4}Institute of Computing Technology, CAS \\
\textsuperscript{5}Peking University \quad
\textsuperscript{6}Australian National University \quad
\textsuperscript{7}City University of Macau \\
{\tt\small xin.shen@uq.edu.au}
}

\begin{document}
% main paper ========================
\maketitle
% \vspace{-2em}
\begin{abstract}
Understanding fine-grained human hand motion is fundamental to visual perception, embodied intelligence, and multimodal communication.
In this work, we propose \textbf{Fine-grained Finger-level Hand Motion Captioning (FingerCap)}, which aims to generate textual descriptions that capture detailed finger-level semantics of hand actions. 
To support this task, we curate \textbf{FingerCap-40K}, a large-scale corpus of 40K paired hand-motion videos and captions spanning two complementary sources: concise instruction-style finger motions and diverse, naturalistic hand–object interactions. 
To enable effective evaluation, we employ \textbf{HandJudge}, a LLM-based rubric that measures finger-level correctness and motion completeness.

Temporal sparsity remains a fundamental bottleneck for current Video-MLLMs, since sparse RGB sampling is insufficient to capture the subtle, high-frequency dynamics underlying fine finger motions. 
As a simple and compute-friendly remedy, we introduce \textbf{FiGOP} (Finger Group-of-Pictures), which pairs each RGB keyframe with subsequent hand keypoints until the next keyframe. 
A lightweight temporal encoder converts the keypoints into motion embeddings and integrates them with RGB features. 
\textit{FiGOP} adapts the classic GOP concept to finger motion, recovering fine temporal cues without increasing RGB density.
Experiments on \textbf{FingerCap-40K} show that strong open- and closed-source Video-MLLMs still struggle with finger-level reasoning, while our FiGOP-augmented model yield consistent gains under \textbf{HandJudge} and human studies. 
We will release the dataset and code upon acceptance.
\end{abstract}
\vspace{-1em}

\section{Introduction}
\label{sec:intro}

\begin{figure}[t]
\begin{center}
   \includegraphics[width=0.90\linewidth]{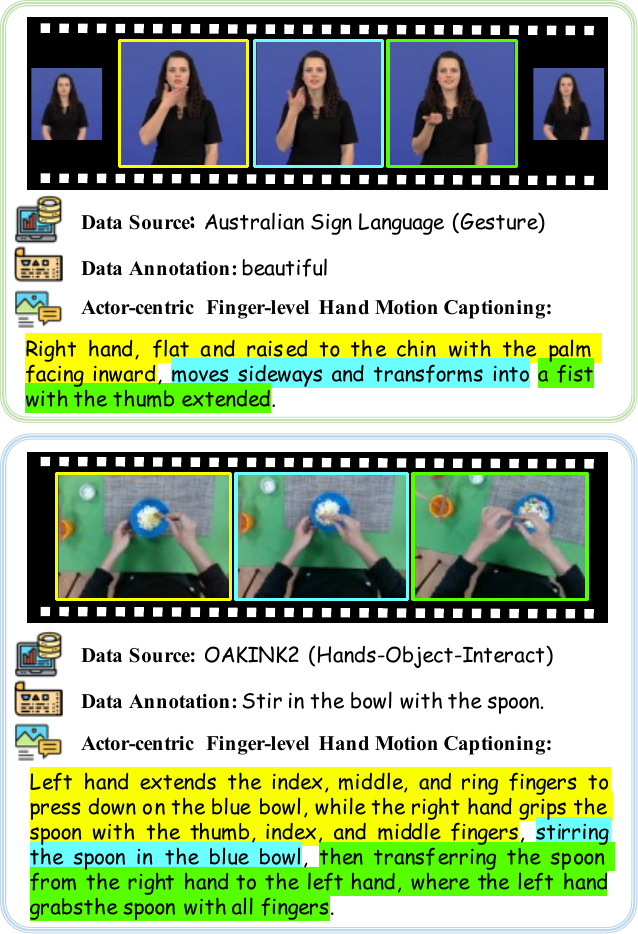}
   \vspace{-0.5em}
   \caption{\textbf{\textit{FingerCap}} aims to generate textual descriptions that capture detailed finger-level semantics of hand actions. 
   Examples from \textbf{\textit{FingerCap-40K}}: top, concise instruction-style clips with explicit targets for finger articulation; bottom, hand--object interactions showing coordinated finger dynamics during manipulation.}
\label{intro_fig}
\end{center}
\vspace{-2.5em}
\end{figure}

Human hand motion is central to both physical manipulation~\cite{GIGAHANDS,HOI4D,OakInk2,DBLP:conf/cvpr/0006LLL025,fan2023arctic} and nonverbal communication~\cite{csl500, WLASL,albanie2020bsl,cao2025socialgesture}. 
From grasping tools and typing~\cite{hoi_gen1,hoi_gen2,hoi_gen3,hoi_gen5} to signing and gesturing~\cite{sign_gen1,sign_gen2,sign_gen3}, the hands convey rich semantic and functional information. 
However, most existing research focuses on coarse hand-level actions~\cite{GIGAHANDS,OakInk2} or global gestures~\cite{MM-WLAUSLAN,cao2025socialgesture}, overlooking the subtle yet crucial contributions of individual fingers. 
These fine-grained finger articulations are essential for dexterity, precision, and intent, and even subtle variations in finger configurations can lead to different gesture semantics or determine whether a manipulation succeeds~\cite{li2025maniptrans,hoi_gen1,hoi_gen2,GIGAHANDS,bilge2022towards}.
To bridge this gap, we introduce \textbf{Fine-grained Finger-level Hand Motion Captioning (\textit{FingerCap})}, a new task that generates detailed textual descriptions of how individual fingers move and coordinate during hand actions.
Unlike conventional motion captioning~\cite{GIGAHANDS,chen2024motionllm,song2025towards,li2025chatmotion,li2025human} or gesture recognition~\cite{WLASL,albanie2020bsl,csl500,MM-WLAUSLAN}, \textit{FingerCap} requires models to capture and describe fine-grained finger articulation, temporal evolution, and inter-finger coordination, whether in communicative gestures or object manipulation.
 
To support the \textit{FingerCap} task, we curate \textbf{\textit{FingerCap-40K}}, a large-scale dataset of 40K video–caption pairs. 
As illustrated in Figure~\ref{intro_fig}, the dataset spans two complementary domains: gesture instruction and hand–object interaction (HOI).
The gesture domain is built from sign language datasets across four regions~\cite{WLASL,albanie2020bsl,csl500,MM-WLAUSLAN}, providing linguistically structured examples of fine finger articulations that are reviewed and refined by sign language experts. 
These samples offer diverse hand configurations and compositional gestures with explicit semantic meaning, serving as high-quality supervision for finger-level understanding~\cite{Auslan_Corpus_new,Auslan_SignBank}.
In contrast, the HOI domain captures physically grounded behaviors such as grasping, twisting, pinching, and transferring objects, collected from large-scale multi-view datasets~\cite{GIGAHANDS,OakInk2} and out-of-distribution benchmarks~\cite{HOI4D,MotionBench}. 
This domain complements the gesture data by introducing natural, unconstrained finger coordination in manipulation tasks.
By integrating linguistically precise gestures with physically diverse interactions, \textit{FingerCap-40K} provides both semantic richness and physical realism, forming a comprehensive foundation for modeling fine-grained finger motion captions.

Current Video-MLLMs~\cite{Qwen-VL,qwen2.5-VL,llavanextvideo,internvl3,internvl3_5,hurst2024gpt,comanici2025gemini,DBLP:conf/emnlp/LinYZCNJ024,qwen3technicalreport} often sample RGB frames sparsely to reduce computation, but this leads to temporal sparsity and fails to capture rapid finger movements. 
To mitigate this issue, we propose \textbf{\textit{FiGOP} (Finger Group-of-Pictures)}, a lightweight and compute-efficient mechanism that augments each sparsely sampled RGB keyframe with the subsequent sequence of 2D hand keypoints~\cite{yang2023effective}, forming a \textit{FiGOP} unit. 
A temporal encoder~\cite{stgcn,vaswani2017attention} then aggregates the keypoint sequence into a compact motion representation that preserves subtle, high-frequency finger articulations, and this representation is integrated with visual tokens within the multimodal projector. 
\textit{FiGOP} adapts the classical GOP~\cite{EMA,jin2024video} concept to hand motion and can be seamlessly applied to existing Video-MLLMs. 
Compared to increasing RGB frame density, \textit{FiGOP} recovers fine-grained temporal cues at significantly lower memory and latency cost while remaining scalable to long sequences.

To enable reliable evaluation of finger-level motion captions, we introduce \textbf{\textit{HandJudge}}, a compact LLM-based assessment framework~\cite{Gu2024AASO, li2025generation,jang2025lost}. 
Conventional captioning metrics~\cite{bleu,lin2004rouge,vedantam2015cider,banerjee2005meteor} are inadequate for \textit{FingerCap} since they fail to capture fine-grained finger articulation and motion dynamics. 
HandJudge evaluates generated captions along four dimensions:
(1) fine-grained finger and hand identification accuracy; 
(2) correctness of finger motion and trajectory; 
(3) fidelity in describing physical interactions with objects; and 
(4) motion coverage, assessing whether the description reflects the full progression of the action.

Through comprehensive experiments on both open- and closed-source Video-MLLMs~\cite{qwen2.5-VL,llavanextvideo,internvl3_5,hurst2024gpt,comanici2025gemini}, as well as task-adapted fine-tuned models, we observe that current models perform poorly on \textit{FingerCap}. 
They frequently miss or misrepresent finger articulations, and in many cases, the generated captions are vague or even hallucinated. 
These results reveal a fundamental gap in fine-grained hand motion understanding and underscore the need for dedicated benchmarks and modeling strategies.

In summary, our contributions are fourfold:
\begin{itemize}
    \item We define \textbf{\textit{FingerCap}}, a new task for understanding and describing detailed finger articulation and coordination.
    \item We curate \textbf{\textit{FingerCap-40K}}, a 40K-sample dataset combining linguistically precise gesture data and physically grounded hand–object interactions.
    \item We introduce \textbf{\textit{FiGOP}}, a compute-efficient module that binds sparse RGB keyframes with dense hand keypoints to capture fine temporal details.
    \item We propose \textbf{\textit{HandJudge}}, an LLM-based evaluation framework for interpretable, multi-dimensional assessment of fine-grained motion semantics.
\end{itemize}

\section{Related Work}
\label{sec:Relatedwork}

\begin{figure*}[t]
\begin{center}
\includegraphics[width=0.93\linewidth]{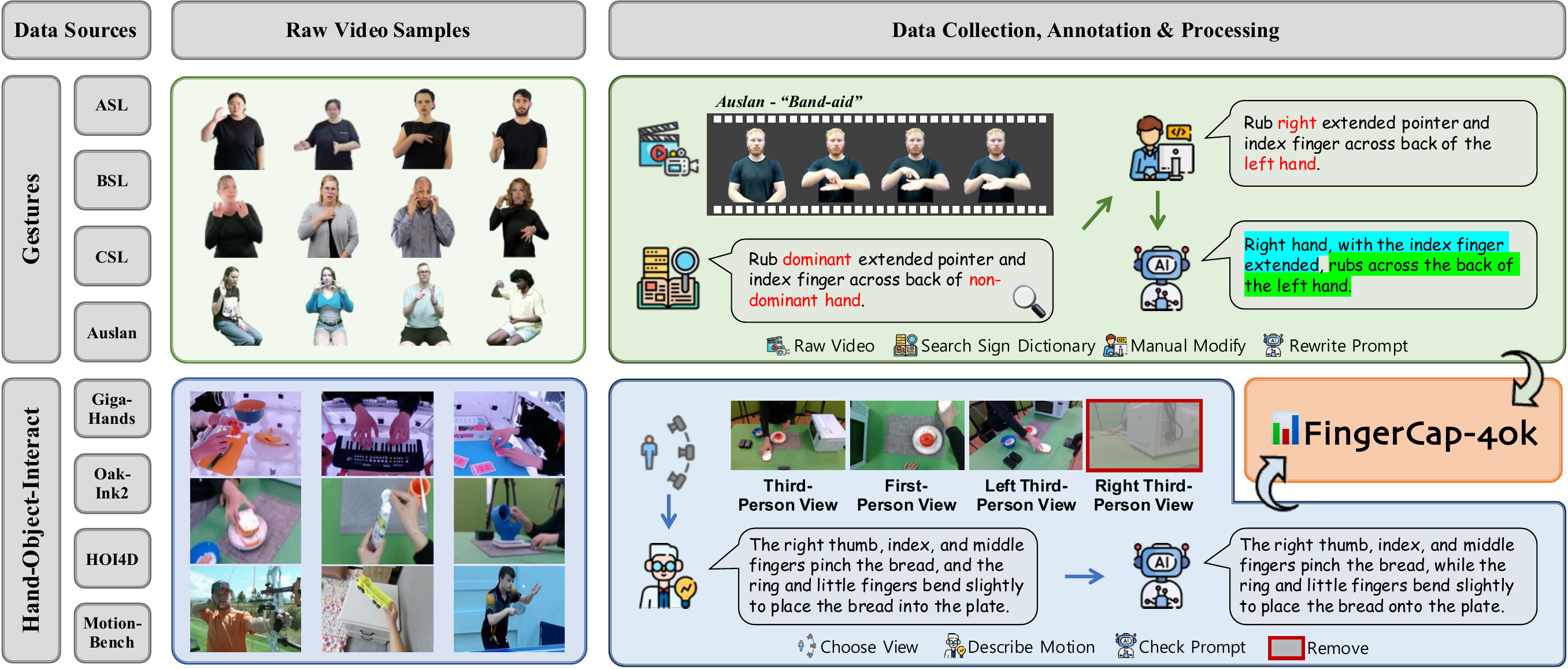}
\end{center}
\vspace{-1.5em}
\caption{Data collection, annotation and processing pipeline for gesture and hand–object interaction data in \textbf{\textit{FingerCap-40K}}.
Gesture videos are collected from multilingual sign language datasets, where raw dictionary-style motion descriptions are manually corrected and refined using an LLM to produce finger-level captions.
Hand–object interaction videos are sampled from multi-view manipulation datasets, in which the clearest view is selected, followed by human-written and LLM-refined finger–object interaction descriptions.
% Both branches produce temporally aligned, natural language annotations with explicit finger articulation.
}
\vspace{-1.5em}
\label{annotation}
\end{figure*} 

\subsection{Hand Motion Datasets}

Research on hand motion understanding has led to the development of diverse datasets across three major domains: hand–object interaction, gesture recognition, and sign language recognition (ISLR). 
Hand–object interaction datasets~\cite{GIGAHANDS,OakInk2,HOI4D,fan2023arctic,liu2024taco,banerjee2024introducing,ohkawa2023assemblyhands,grauman2024ego} have enabled progress in modeling physical manipulation and contact dynamics, supporting studies of grasping and coordination~\cite{hoi_gen1,hoi_gen2,hoi_gen3}.
However, these datasets primarily focus on coarse action categories or hand trajectories, without offering detailed representations of finger articulation or semantic descriptions that capture intent.
Gesture recognition and ISLR datasets, in contrast, focus on communicative and symbolic hand movements. 
While they vary in scale from dozens to thousands of gesture or gloss classes, they often remain limited to single-view RGB recordings, restricted viewpoints, or small vocabularies~\cite{csl500,starner2023popsign,cao2025socialgesture,liu2022ld,liu2021imigue,perera2018uav}.
Although recent datasets have improved realism through larger signer diversity and higher resolution, they typically describe isolated gestures rather than continuous, fine-grained finger motion~\cite{albanie2020bsl,MM-WLAUSLAN,WLASL,joze2018ms}.
This limitation hinders the ability of these datasets to represent the subtle articulations and temporal coordination that underlie expressive and dexterous hand behavior~\cite{bilge2022towards}.
Existing hand motion datasets have greatly advanced recognition and interaction understanding, but they largely neglect continuous finger-level dynamics and lack natural language grounding. 
To address this gap, our \textit{FingerCap-40K} dataset provides large-scale paired video–text data covering both structured gestures and natural hand–object interactions, enabling a new research direction in fine-grained finger motion captioning.

\subsection{Human Motion Understanding}
Traditional motion understanding methods are predominantly built on 3D skeletal representations, where actions are modeled as sequences of articulated joints~\cite{zhu2025motiongpt3,jiang2023motiongpt,guo2022tm2t,li2024lamp,wu2024mote}. 
These datasets and models are typically annotated at the body or hand level, without explicit supervision for individual fingers. 
As a result, they can capture coarse motion patterns but are fundamentally unable to learn how specific fingers articulate, coordinate, or interact with objects. 
Even hand-centric datasets and models~\cite{GIGAHANDS,OakInk2,HOI4D} follow this design and describe hand pose at a single rigid unit, rather than providing finger-level trajectories.
To enrich geometric modeling with visual cues, recent works combine RGB video and pose sequences~\cite{song2025towards,chen2024motionllm,li2025human,hwang2025motif,li2025chatmotion}. 
However, the underlying annotations still operate at the hand level, and pose streams are usually sampled at low temporal resolution. 
Consequently, high-frequency finger movements and subtle transitions between finger configurations are either missing from the data or smoothed out during aggregation, preventing these models from developing true finger-level understanding.

Video Multimodal Large Language Models (Video-MLLMs)~\cite{llavanextvideo,qwen2.5-VL,internvl3,internvl3_5,hurst2024gpt,comanici2025gemini} introduce powerful language reasoning on top of visual features and show emerging ability to describe hand actions from large-scale web data. 
However, they inherit the same limitations of their training corpora: sparse frame sampling and predominantly hand-level supervision. 
They can often infer the overall intent of a gesture, but they rarely capture which fingers move, in what order, and how they make or break contact.
To efficiently mitigate temporal sparsity and expose models to explicit finger supervision, we propose \textbf{\textit{FiGOP}}, which enriches Video-MLLMs with fine-grained hand keypoints for finger-level understanding.

\begin{figure*}[t]
    \centering

    % 左侧：真正的 Table 1
    \begin{minipage}[t]{0.45\linewidth}
        \captionsetup{type=table}
        \centering
        % \caption{Statistics of the \textbf{\textit{FingerCap-40K}} dataset. 
        % The gesture subset is collected from multilingual sign language corpora, while the hand–object interaction subset consists of natural manipulation sequences. 
        % We report the number of videos, total frames, caption words, vocabulary size, camera views, and the distribution of single-hand and double-hand motions.}
        \caption{
        Statistics of the \textbf{\textit{FingerCap-40K}} dataset across gesture and hand–object interaction domains, summarizing video and text scale, frame density, vocabulary coverage, camera diversity, hand-use distribution, and OOD subsets information.
        }
        \label{tab:dataset_stats}
        \footnotesize
        \renewcommand{\arraystretch}{1.25}
        \begin{tabular}{l|cc}
            \toprule
            & Gesture & Hand--Object Interaction \\
            \midrule
            Data Source   & ASL, CSL, Auslan & GigaHands, OakInk2 \\
            Num.Videos    & 21,055 & 19,922 \\
            Num.Words     & 651,112 & 744,296 \\
            Num.Frames    & 1,922,471 & 4,251,389 \\
            Num.Vocs      & 3,427 & 4,882 \\
            Num.Views     & 3 & 5 \\
            Single-Hand   & 6,850 & 1,544 \\
            Both-Hand     & 14,205 & 18,378 \\
            \midrule
            OOD Set       & BSL & HOI4D, MotionBench \\
            OOD Vocs      & 100 & 36 \\
            OOD Domain    & sign language & sport, medical, ... \\
            \bottomrule
        \end{tabular}
    \end{minipage}
    \hfill
    % 右侧：真正的 Figure
    \begin{minipage}[t]{0.52\linewidth}
    % \caption{
    % Data distribution in \textbf{\textit{FingerCap-40K}}.
    % Top: word cloud of finger- and hand-related terms in the captions, highlighting frequent references. 
    % Bottom left: distribution of video durations for the train, validation and test splits. 
    % Bottom middle: ratio of single-hand and double-hand motions for each camera viewpoint. 
    % Bottom right: distribution of caption lengths across the three splits.
    % }
    \vspace{-0.2em}
        \captionsetup{type=figure}
        \centering
        \includegraphics[width=\linewidth]{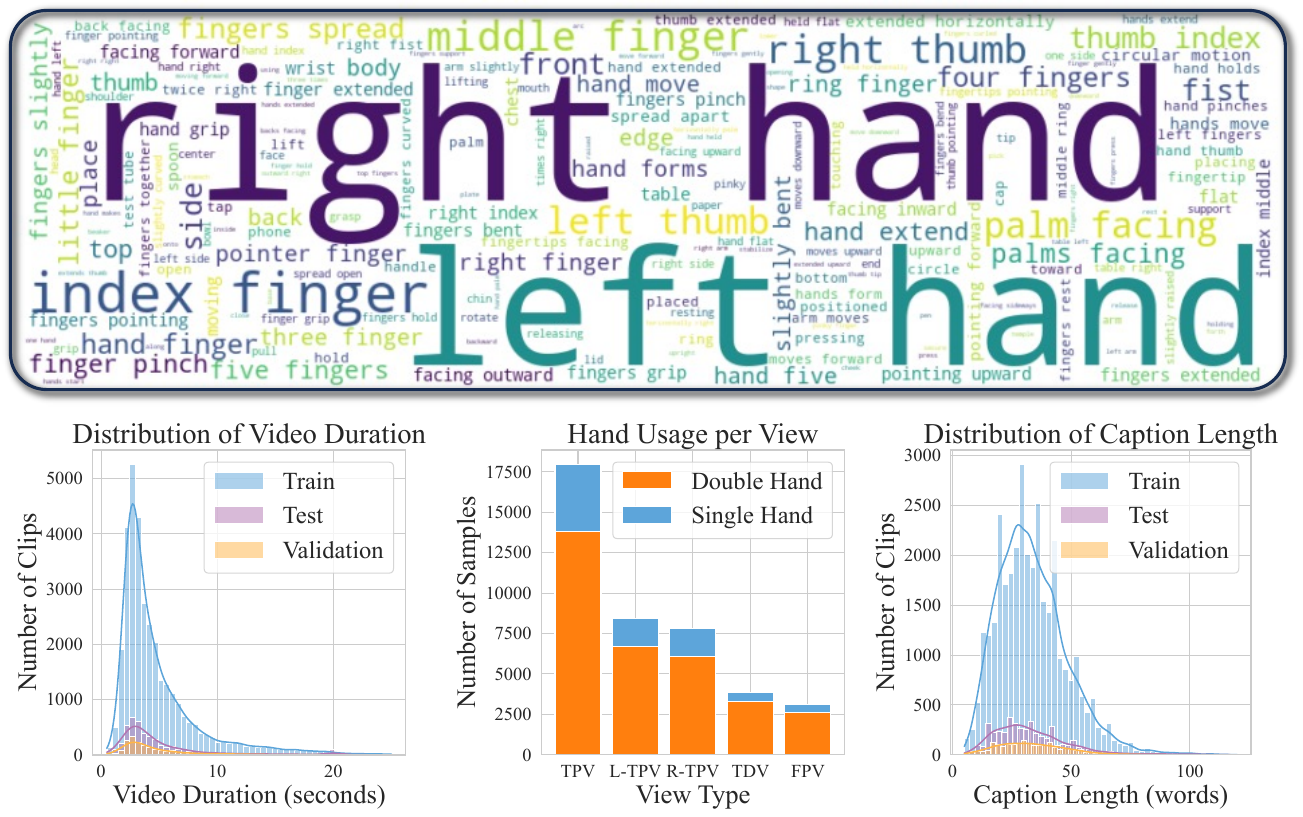}
    \caption{
    Data distribution in \textit{\textbf{FingerCap-40K}}.
    Top: the word cloud of finger- and hand-related terms in captions.
    Bottom: (left) video duration; (middle) single vs. double hand usage across viewpoints\protect\footnotemark; and (right) caption length distribution.
    }
    \label{fig:dataset_dist}
    \end{minipage}
    
\vspace{-1.5em}
\end{figure*}

\section{FingerCap-40K} 
\label{sec:dataset}

Building reliable models for finger-level hand motion understanding requires data that jointly capture both \textit{semantic intent} and \textit{fine-grained kinematics}.
However, existing datasets fall short in several aspects: 
(1) gesture corpora~\cite{MM-WLAUSLAN,cao2025socialgesture} often contain high-level linguistic semantics but lack explicit descriptions of finger articulations; 
(2) hand–object datasets~\cite{grauman2024ego,fan2023arctic} focus on manipulation dynamics yet rarely include natural language annotations that describe motion in finger-level detail; 
and (3) few resources~\cite{GIGAHANDS,OakInk2} provide temporally aligned motion–caption pairs with consistent left–right and contact annotations.
These limitations hinder the study of fine-grained motion reasoning and the evaluation of multimodal models under precise physical supervision.
To bridge this gap, we curate \textbf{\textit{FingerCap-40K}}\footnote{\textit{FingerCap-40K} and all sources follow the \textbf{CC BY-NC-SA 4.0} license.}, a large-scale video–caption dataset featuring 40K fine-grained hand motion–language pairs. 

\subsection{Data Sources}

\textbf{\textit{FingerCap-40K}} is constructed from two complementary domains: gesture instruction and hand–object interaction. 
The gesture subset is collected from four major sign language systems, including ASL~\cite{WLASL}, BSL~\cite{albanie2020bsl}, CSL~\cite{csl500} and Auslan~\cite{MM-WLAUSLAN}. 
These videos provide naturally aligned motion–text pairs with semantically precise finger articulations. 
All annotations are reviewed and refined with the assistance of sign language experts to ensure structural correctness and temporal consistency.
The hand–object interaction subset captures physically grounded finger motions such as grasping, twisting and pinching, sampled from large-scale multi-view datasets such as GigaHands~\cite{GIGAHANDS} and OakInk2~\cite{OakInk2}.
To evaluate generalization under distribution shifts, we further include out-of-distribution samples from HOI4D~\cite{HOI4D} and MotionBench~\cite{MotionBench}. 
Together, these two domains provide both semantic precision and kinematic diversity, forming a comprehensive foundation for fine-grained finger-level hand motion captioning.

\subsection{Data Collection, Annotation and Processing}

Based on the two domains introduced above, we construct \textit{FingerCap-40K} through a unified pipeline (Figure~\ref{annotation}) that ensures temporal alignment, semantic precision and natural language quality in all video–caption pairs.

\textbf{Gesture data.}
For sign language videos, we first retrieve motion descriptions from official sign language dictionaries corresponding to each dataset. 
These raw descriptions are not directly suitable for captioning due to three issues: 
(1) hand references are often written as ``dominant'' and ``non-dominant'' rather than left and right, which creates spatial ambiguity; 
(2) some entries include non-motor content such as emotions or analogies unrelated to physical motion; and
(3) the language is instructional and lacks the natural sentence structure expected in captions. 
To address this, we manually correct or remove ambiguous content, normalize left and right hand references, and then paraphrase the text using GPT-4.1~\cite{achiam2023gpt} while preserving the original motion semantics. 
The resulting captions are concise, fluent, and finger-level accurate.

\textbf{Hand–object interaction (HOI) data.}
For HOI videos, we sample clips from multiple viewpoints, including first-person, third-person, left, right and top-down cameras, and only retain those in which all finger movements are clearly visible. 
Annotators then describe how each finger interacts with the object, focusing on articulation, contact and coordination between both hands when present. 
These descriptions are subsequently refined using GPT-4.1~\cite{achiam2023gpt} to ensure grammatical consistency and descriptive clarity.

% \textbf{Out-of-distribution data.}
% To evaluate generalization, we include British Sign Language (BSL), HOI4D~\cite{HOI4D} and MotionBench~\cite{MotionBench} in the test set. These samples introduce unseen languages, objects and task scenarios, enabling assessment of robustness under linguistic and physical distribution shifts.

% Through this multi-stage pipeline, \textit{FingerCap-40K} offers finger-level motion captions that are temporally aligned, semantically detailed and linguistically natural across both communicative and manipulative contexts.

\subsection{Dataset Statistics}

Table~\ref{tab:dataset_stats} presents the core statistics of \textit{FingerCap-40K}, highlighting its scale and diversity. 
It contains 40K video with fine-grained finger-level hand motion caption, including 21K gesture clips and 19K hand–object interaction clips. 
In total, it comprises 6.17 million frames and 1.40 million caption words. 
The gesture subset has a vocabulary size of 3.4K, and the interaction subset has 4.8K unique words, indicating substantial linguistic diversity across communicative and manipulative domains.
For training and evaluation, the dataset is split into training, validation, and test sets in a ratio of 8:1:1.

\footnotetext{
TPV = third-person view, L-TPV = left third-person view, R-TPV = right third-person view,
TDV = top-down view, FPV = first-person view.}

Figure~\ref{fig:dataset_dist} further analyzes the data distribution. 
The word cloud (top) shows frequent use of terms referring to individual fingers and hands, which reflects the dataset focus on fine-grained motion semantics.
The video durations (bottom-left) are mostly between one and ten seconds, producing short but motion-rich clips. 
Caption lengths (bottom-right) exhibit a long-tailed distribution, indicating varied levels of descriptive complexity across samples.
Hand usage across TPV, L-TPV, R-TPV, TDV and FPV (bottom-middle) shows that bimanual actions occur more often than single-hand motions, offering a wide range of viewpoints that support modeling detailed finger movements under varied conditions.

\section{\textit{FiGOP}-augmented Video-MLLM}
\label{sec:FiGOP}

\begin{figure}[t]
\begin{center}
   \includegraphics[width=1\linewidth]{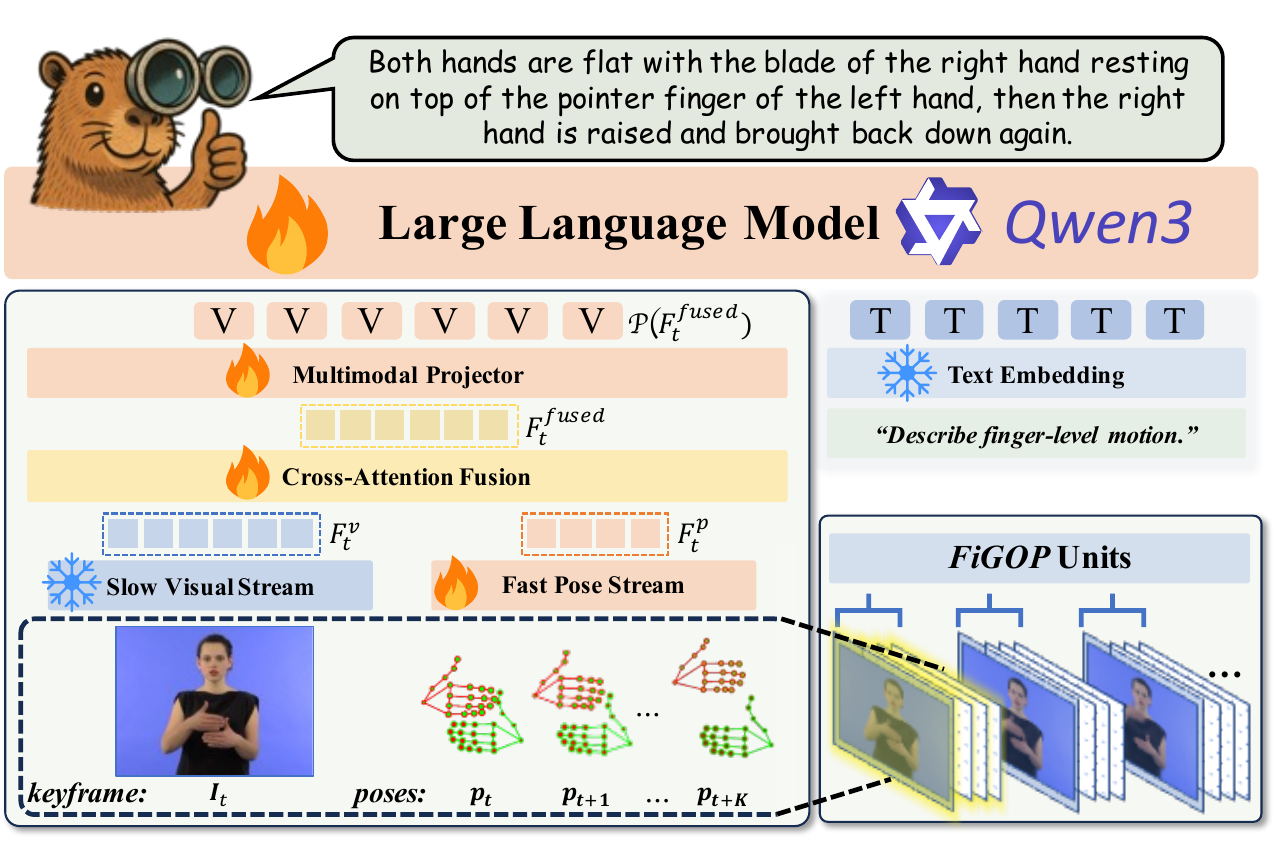}
   \vspace{-1.5em}
   \caption{
   Overview of the \textit{FiGOP}-augmented Video-MLLM.
   % combining RGB frames and hand poses for enhanced finger-level motion understanding.
   }
\label{figop}
\end{center}
\vspace{-2.5em}
\end{figure}

Current Video-MLLMs~\cite{Qwen-VL,qwen2.5-VL,llavanextvideo,internvl3,internvl3_5,hurst2024gpt,comanici2025gemini,DBLP:conf/emnlp/LinYZCNJ024,qwen3technicalreport} typically adopt sparse RGB sampling to reduce computational cost when processing long videos.
However, such sampling discards high-frequency motion cues, especially the rapid articulations and coordination of fingers. 
Inspired by the Group-of-Pictures (GOP) principle~\cite{EMA,jin2024video}, we introduce \textbf{Finger Group-of-Pictures (\textit{FiGOP})}, a simple yet effective encoding mechanism that augments sparsely sampled RGB frames with dense hand pose streams. 
Unlike prior work that relies on optical flow or dense frame inputs~\cite{EMA,jin2024video}, \textit{FiGOP} uses structurally organized 2D hand keypoints~\cite{yang2023effective}, which provide explicit motion trajectories without increasing pixel-level redundancy.
An overview of this architecture is shown in Figure~\ref{figop}.

\subsection{\textit{\textbf{FiGOP}} Unit Construction}
A video is divided into a sequence of \textit{FiGOP} units. 
Each unit consists of:
\begin{equation}
\text{\textit{FiGOP}}_t = \left( I_t, \, P_{t:t+K} \right),
\end{equation}
where $I_t \in \mathbb{R}^{H \times W \times 3}$ is a sparsely sampled RGB keyframe, and $P_{t:t+K} = \{p_t, p_{t+1}, \dots, p_{t+K-1}\}$ is a dense sequence of hand poses between the current and next keyframe. 
Each $p_i \in \mathbb{R}^{J \times C}$ represents $J$ hand joints, with $C$-dimensional features (\eg, $(x,y)$ coordinates and confidence)~\cite{yang2023effective}. 
This design preserves high-frequency motion while keeping RGB sampling unchanged.

\subsection{Dual-Stream Encoding}
Each \textit{FiGOP} unit is processed by two parallel streams:
\begin{itemize}
\item \textbf{Slow Visual Stream.} 
The RGB keyframe $I_t$ is encoded by a pre-trained vision encoder~\cite{dosovitskiy2020image,qwen3technicalreport,qwen2.5-VL} to obtain spatial tokens $F_t^v \in \mathbb{R}^{N \times D_v}$.

\item \textbf{Fast Pose Stream.} 
The pose sequence $P_{t:t+K}$ is fed into the ST-GCN module~\cite{stgcn,cosign, shen2025cross} to model finger joint topology and local motion. 
A lightweight temporal Transformer~\cite{vaswani2017attention} further aggregates cross-frame dependencies, producing $F_t^p \in \mathbb{R}^{K \times D_p}$.
\end{itemize}

\noindent Pose representations, unlike optical flow or RGB images, are structured and physically meaningful, offering computational efficiency and robustness to background and lighting variations~\cite{cosign,shen2025cross}.

\subsection{Motion-Aware Projector Fusion}
To inject high-frequency motion cues into the visual representation, we incorporate a motion-aware adapter~\cite{EMA} into the multimodal projector. 
Given visual tokens $F_t^v$ and pose motion features $F_t^p$, we apply a cross-attention~\cite{vaswani2017attention} fusion:
\begin{equation}
F_t^{fused} = \text{CA}(F_t^v, F_t^p) + F_t^v,
\end{equation}
\begin{equation}
\text{CA}(F_t^v, F_t^p) = \text{Attn}(F_t^v W_Q,\; F_t^p W_K,\; F_t^p W_V),
\end{equation}
\noindent where $W_Q, W_K$ and $W_V$ denote learnable projection matrices.
The fused tokens are then projected into the LLM (\textbf{\textit{Qwen3}}~\cite{qwen3technicalreport}) embedding space:

\begin{equation}
E_t^{LLM} = \mathcal{P}(F_t^{fused}).
\end{equation}

\noindent Our design allows visual tokens to selectively retrieve motion information from pose features, improving finger-level temporal reasoning.

\section{Experiments}
\label{sec:Experiments}

% \subsection{Task Definition}

% The Fine-grained Finger-level Hand Motion Captioning (\textit{FingerCap}) task aims to generate natural-language descriptions that explicitly capture how individual fingers articulate and coordinate during a hand motion sequence.
% Given a video $V = \{f_1, f_2, \dots, f_T\}$ containing a gesture or hand--object interaction, the objective is to produce a caption 
% $C = \{w_1, w_2, \dots, w_N\}$ that specifies: 
% (1) which finger(s) and hand(s) are involved; 
% (2) how they move and evolve over time; and 
% (3) how they interact with objects, the other hand, or other body parts.
% Unlike general action captioning or gesture recognition, which focuses on global intent, this task requires anatomical precision, temporal coherence, and contact-aware reasoning at the finger level.

\begin{table*}[ht]
\centering
\footnotesize
\renewcommand{\arraystretch}{1.2}
\setlength{\tabcolsep}{4.5pt}
\caption{Performance comparison of different models on standard captioning metrics in the \textit{\textbf{FingerCap-40K}} test set.}
\label{tab:caption_result}
\begin{tabular}{l|cccc|cccc|cccc}
\toprule
\multirow{2}{*}{\textbf{Model}} & 
\multicolumn{4}{c|}{\textbf{Gesture}} & 
\multicolumn{4}{c|}{\textbf{HOI}} & 
\multicolumn{4}{c}{\textbf{Average}} \\
\cmidrule(lr){2-5} \cmidrule(lr){6-9} \cmidrule(lr){10-13}
 & B-4 & R-L & METEOR & CIDEr  & B-4 & R-L & METEOR & CIDEr  & B-4 & R-L & METEOR & CIDEr  \\
\midrule
\multicolumn{10}{l}{\textbf{\textit{Close-source Models}}} \\
\midrule
GPT-4o~\cite{hurst2024gpt} &1.54&18.36&27.95&24.64 &2.71&22.12&28.42&17.54&2.09&20.13&28.17&21.29\\
GPT-4o-mini~\cite{hurst2024gpt} & 1.11&14.15&25.58&21.05 &1.60&13.66&26.15&13.06&1.34&13.92&25.85&17.28\\
Gemini-2.5-Pro~\cite{comanici2025gemini} & 3.31&27.13&35.13&29.69&3.30&24.46&33.33&35.05&3.31&25.87&34.28&32.37\\
\midrule
\multicolumn{10}{l}{\textbf{\textit{Open-source Models}}} \\
\midrule
LLaVA-NeXT-Video-7B~\cite{llavanextvideo} &  1.41&19.32&27.02&13.85&2.78&21.73&28.23&11.63&2.05&20.46&27.59&12.80  \\
InternVL3-8B~\cite{internvl3} & 1.33&21.49&28.98&18.38&2.45&23.72&29.87&28.76&1.86&22.54&29.40&23.28\\
InternVL3.5-8B~\cite{internvl3_5} & 0.93&19.97&27.91&16.45&2.33&22.82&30.15&17.26&1.59&21.31&28.97&16.83\\
Qwen2.5-VL-7B-Instruct~\cite{qwen2.5-VL} & 1.24& 20.67&26.15&23.84&2.58&23.10&27.77&34.91&1.87&21.81&26.92&29.06  \\
Qwen3-VL-8B-Instruct~\cite{qwen3technicalreport} &1.98&19.85&30.30&23.10&2.79&19.27&30.01&35.92&2.36&19.58&30.16&29.14  \\
\midrule
\multicolumn{10}{l}{\textbf{\textit{Fine-tuned Models: Qwen3-VL-8B-Instruct}}~\cite{qwen3technicalreport}} \\
\midrule
+ \textit{MM Projector} + SFT  & 7.84&31.17&32.87&89.89&13.42&36.15&36.36&126.73&10.48&33.52&34.51&107.28  \\
% InternVL3.5-8B &  \textbf{16.99}&\textbf{39.47}&\textbf{40.94}&\textbf{172.81}&\textbf{20.61}&\textbf{41.91}&\textbf{42.42}&\textbf{198.70}&\textbf{18.70}&\textbf{40.62}&\textbf{41.64}&\textbf{185.03}  \\
\rowcolor[HTML]{EFEFEF}  + \textbf{\textit{FiGOP}} + SFT (ours)  &  \textbf{13.81}&\textbf{36.84}&\textbf{38.41}&\textbf{146.29}&\textbf{17.09}&\textbf{39.14}&\textbf{39.43}&\textbf{165.31}&\textbf{15.36}&\textbf{37.92}&\textbf{38.89}&\textbf{155.27}\\
\bottomrule
\end{tabular}
\vspace{-1.5em}
\end{table*}

\subsection{Evaluation Metrics}

\noindent\textbf{Standard Caption Metrics.}
Following common practice in video captioning, we report BLEU-4 (\textbf{B-4})~\cite{bleu}, ROUGE-L (\textbf{R-L})~\cite{lin2004rouge}, METEOR (\textbf{M})~\cite{banerjee2005meteor}, and CIDEr (\textbf{C})~\cite{vedantam2015cider} (all values are multiplied by 100 for clearer comparison after using the \textit{Jury} evaluation toolkit~\cite{cavusoglu2023jury}). 
These metrics measure lexical overlap and fluency but do not effectively capture finger-level correctness, motion direction, or contact semantics, which are crucial for this task.

\noindent\textbf{\textit{HandJudge} (LLM-as-a-Judge).}
To address the limitations of standard metrics, we introduce \textit{HandJudge}, an LLM-based evaluation framework~\cite{Gu2024AASO, li2025generation,jang2025lost} using GPT-4.1~\cite{achiam2023gpt}. 
For each generated caption, the LLM compares it with the ground truth and assigns a score from 0 to 5 across four expert-defined criteria:
(1) finger and hand identification (\textbf{FHI}), 
(2) motion and trajectory accuracy (\textbf{MT}), 
(3) contact and interaction reasoning (\textbf{CI}), and 
(4) completeness of motion sequence (\textbf{CMS}). 
The LLM also provides intermediate reasoning before scoring, offering interpretable and fine-grained assessment that better aligns with human judgment, as illustrated in Figure~\ref{case_study}. 
More details are provided in the \textit{Appendix}.

\subsection{Implementation Details}
\noindent\textbf{Evaluation Protocol.}
We evaluate several open-source Video-MLLMs using LLaMA-Factory~\cite{llamafactory}. 
% and their official checkpoints.
Closed-source systems are queried via APIs with unified decoding settings.
All models receive the same 2-fps-sampled RGB inputs, ensuring a fair comparison. 

\noindent\textbf{\textit{FiGOP}-augmented Video-MLLM.}
We apply Qwen3-VL-8B~\cite{qwen2.5-VL,qwen3technicalreport} as the backbone, and use the efficient and precise DWPose~\cite{yang2023effective} to extract 2D poses. 
% with 42 keypoints from both hands.
Videos are sampled at 2 fps, and for each RGB keyframe, we attach a 2D hand-pose sequence in the following 8 frames, forming a \textit{FiGOP} unit.
RGB frames are encoded by the frozen vision tower~\cite{qwen3technicalreport,dosovitskiy2020image}, while poses are processed through a two-layer ST-GCN and temporal Transformer to generate pose motion embeddings.

\noindent\textbf{Two-Stage Fine-tuning.}
We perform full supervised fine-tuning on \textit{FingerCap-40K}. 
In Stage 1, we freeze the vision encoder and the LLM, and train the pose encoder and the projector for one epoch. 
In Stage 2, we unfreeze the LLM and fine-tune both projector and LLM with a next-token-prediction loss for three epochs.
The whole process is trained on eight NVIDIA A100 GPUs with a batch size of 1 and learning rates of $1\mathrm{e}{-4}$ (Stage 1) and $1\mathrm{e}{-5}$ (Stage 2).
\noindent More details of prompts, experimental settings, and evaluation are provided in the \textit{Appendix}.

\subsection{Baseline Models}

\noindent\textbf{Closed-source Models.}  
We evaluate three state-of-the-art proprietary multimodal systems: \textit{GPT-4o}~\cite{hurst2024gpt}, \textit{GPT-4o-mini}~\cite{hurst2024gpt}, and \textit{Gemini-2.5-Pro}~\cite{comanici2025gemini}. 
These models are capable of understanding videos and represent the frontier in commercial multimodal reasoning.

\noindent\textbf{Open-source Models.}  
We include several recent state-of-the-art open-source MLLMs: \textit{LLaVA-NeXT-Video-7B}~\cite{llavanextvideo}, \textit{InternVL3-8B}~\cite{internvl3}, \textit{InternVL3.5-8B}~\cite{internvl3_5}, \textit{Qwen2.5-VL-7B-Instruct}~\cite{qwen2.5-VL}, and \textit{Qwen3-VL-8B-Instruct}~\cite{qwen3technicalreport}. 
These models, trained on large-scale visual–textual alignment and temporal adaptation, serve as transparent and reproducible baselines for fine-grained motion understanding.

\noindent\textbf{Fine-tuned Variants.}  
To evaluate task adaptation, we fine-tune \textit{Qwen3-VL-8B-Instruct} on the \textit{FingerCap-40K} dataset under two configurations: (1) using a standard multimodal projector (\textit{MM Projector}), and (2) using our proposed \textit{FiGOP}-augmented projector, which incorporates structured pose representations through the spatial–temporal fusion.

\begin{table*}[ht]
\centering
\footnotesize
\renewcommand{\arraystretch}{1.2}
\setlength{\tabcolsep}{7pt}
\caption{Results of \textit{\textbf{HandJudge}} evaluation across four dimensions: Finger and Hand Identification (FHI), Motion and Trajectory (MT), Contact and Interaction (CI), and Completeness of Motion Sequence (CMS).}
\label{llm_as_a_judge}
\begin{tabular}{l|cccc|cccc|cccc|c}
\toprule
\multirow{2}{*}{\textbf{Model}} & 
\multicolumn{4}{c|}{\textbf{Gesture}} & 
\multicolumn{4}{c|}{\textbf{HOI}} & 
\multicolumn{5}{c}{\textbf{Average}} \\
\cmidrule(lr){2-5} \cmidrule(lr){6-9} \cmidrule(lr){10-14}
 & FHI & MT & CI & CMS & FHI & MT & CI & CMS & FHI & MT & CI & CMS & Overall \\
\midrule
\multicolumn{10}{l}{\textbf{\textit{Close-source Models}}} \\
\midrule
GPT-4o & 2.26&1.75&1.28&2.94&2.86&2.34&2.17&2.65&2.54&2.03&1.70&2.81&2.27  \\
GPT-4o-mini & 1.79&1.36&1.20&2.26&2.17&1.95&1.81&2.40&1.97&1.64&1.49&2.33&1.86  \\
Gemini-2.5-Pro~\cite{comanici2025gemini} & 2.91&2.13&1.97&\textbf{3.38}&2.53& 2.45&2.40&\textbf{3.33}&2.73&2.28&2.17&\textbf{3.36}&2.64  \\
\midrule
\multicolumn{10}{l}{\textbf{\textit{Open-source Models}}} \\
\midrule
LLaVA-NeXT-Video-7B~\cite{llavanextvideo} & 1.31&1.13&0.89&1.97&1.59&1.21&1.14&1.87&1.44&1.17&1.01&1.92&1.39  \\
InternVL3-8B~\cite{internvl3} & 2.02&1.66&1.27&2.68&2.01&1.92&1.72&2.39&2.01&1.78&1.48&2.54&1.96  \\
InternVL3.5-8B~\cite{internvl3_5} & 2.05&1.59&1.24&2.86&2.24&1.99&1.84&2.73&2.14&1.78&1.52&2.80&2.06  \\
Qwen2.5-VL-7B-Instruct~\cite{qwen2.5-VL} & 1.69&1.30&0.78&1.86&1.92&1.67&1.56&2.09&1.80&1.48&1.15&1.96&1.60  \\
Qwen3-VL-8B-Instruct~\cite{qwen3technicalreport} & 2.14 &1.47  &1.16  &  2.67& 2.64 & 1.93 &1.92  & 2.92 & 2.38&1.69 & 1.52& 2.79&2.09 \\
\midrule
\multicolumn{10}{l}{\textbf{\textit{Fine-tuned Models: Qwen3-VL-8B-Instruct}}~\cite{qwen3technicalreport}} \\
\midrule
+ \textit{MM Projector} + SFT  & 2.40&1.75&1.63&2.50&2.45&1.94&1.99&2.43&2.42&1.84&1.80&2.47&2.13  \\
\rowcolor[HTML]{EFEFEF} + \textbf{\textit{FiGOP}} + SFT (ours) & \textbf{3.03} & \textbf{2.41} & \textbf{2.34} &3.11  & \textbf{2.88} & \textbf{2.60} & \textbf{2.58} & 3.01 & \textbf{2.96}& \textbf{2.50}&\textbf{2.45} &3.06&\textbf{2.74} \\
\bottomrule
\end{tabular}
\vspace{-1.5em}
\end{table*}

\subsection{Results}

% Table~\ref{tab:caption_result} and Table~\ref{llm_as_a_judge} present the performance of various models in the \textit{FingerCap-40K} test set, evaluated using both standard captioning metrics and the proposed \textit{HandJudge}.

% ===========================================================
\noindent \textbf{Zero-shot Evaluation: Closed-source vs. Open-source Models.}  
We first compare the performance of closed-source and open-source models in both standard captioning metrics and \textit{HandJudge} evaluation. As shown in Table~\ref{tab:caption_result} and Table~\ref{llm_as_a_judge}, \textit{Gemini-2.5-Pro}~\cite{comanici2025gemini} outperforms all other models in both evaluation settings. 
In the standard metrics, it achieves the highest METEOR (34.28) and CIDEr (32.37) scores.
Meanwhile, \textit{Gemini-2.5-Pro} maintains a leading position in \textit{HandJudge} with an overall score of 2.64, showing superior performance in FHI and CMS.
In comparison, the open-source models, especially \textit{Qwen3-VL-8B-Instruct}~\cite{qwen3technicalreport}, show competitive performance but fall short of \textit{Gemini-2.5-Pro}.  
Specifically, though \textit{Qwen3-VL-8B-Instruct} achieves a decent average CIDEr score of 29.14, it still lags behind \textit{Gemini-2.5-Pro} in \textit{HandJudge}, with an overall score of 2.09 compared to 2.64 for \textit{Gemini-2.5-Pro}. 
This clearly indicates that closed-source models~\cite{qwen3technicalreport,internvl3_5,llavanextvideo} perform better at capturing both lexical fluency and fine-grained hand motion details in comparison to their open-source~\cite{comanici2025gemini,hurst2024gpt} counterparts.

\noindent \textbf{Impact of Fine-tuning.}  
In the standard captioning evaluation, the fine-tuned \textit{Qwen3-VL-8B-Instruct} demonstrates significant improvements across all metrics. 
Meanwhile, fine-tuning with the standard multimodal projector (\textit{MM Projector}) also results in a substantial increase in performance, and even surpasses \textit{Gemini-2.5-Pro}. 
This indicates that task-specific fine-tuning can boost the model's ability to generate accurate captions.
However, in the \textit{HandJudge} evaluation, the fine-tuned model with \textit{MM Projector} still lags behind \textit{Gemini-2.5-Pro}. 
The model achieves an overall \textit{HandJudge} score of 2.13 compared with 2.64 achieved by \textit{Gemini-2.5-Pro}. 
This suggests that while fine-tuning has a significant impact on caption generation, the finger-level accuracy remains a challenge, highlighting the importance of incorporating rich and task-specific data for fine-grained motion understanding.

\noindent \textbf{\textit{FiGOP}-augmented Video-MLLM.} 
The introduction of \textit{FiGOP} significantly enhances the performance of fine-tuned model. 
The \textit{FiGOP}-augmented model shows improvements in both HOI and gesture subset, surpassing Gemini-2.5-Pro in multiple \textit{HandJudge} metrics. 
Specifically, FHI, MT, and CI scores reach 2.96, 2.50 and 2.45, respectively, and the overall \textit{HandJudge} score increases to 2.74, bringing the performance of the open-source model on par with the closed-source model in several aspects. 
These results suggest that \textit{FiGOP} better captures fine-grained finger motion, mitigating limitations observed in earlier fine-tuned models and narrowing the performance gap between open-source and closed-source models.

% ===========================================================
\begin{figure*}[ht]
\begin{center}
\includegraphics[width=0.90\linewidth]{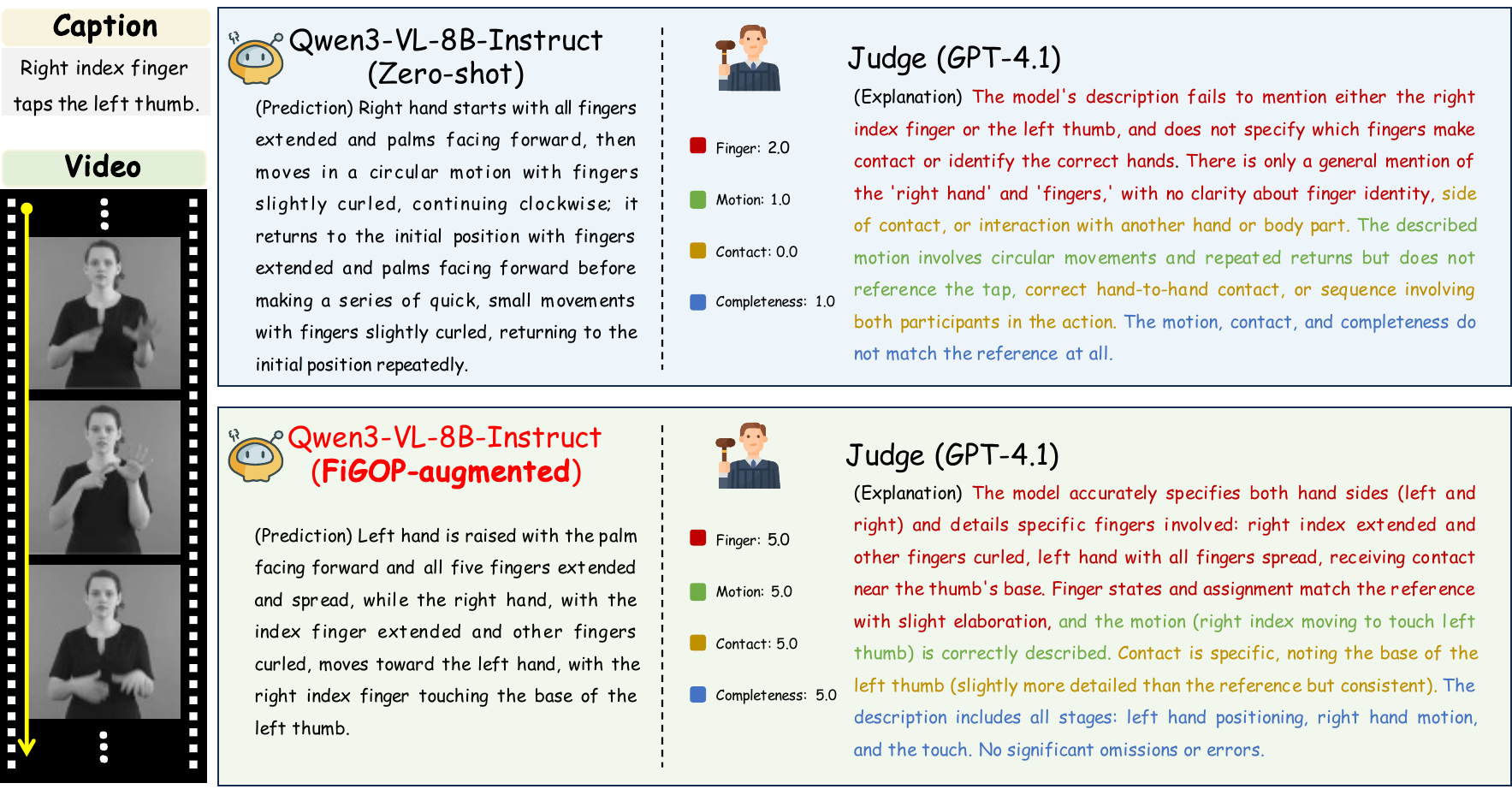}
\end{center}
\vspace{-2em}
\caption{Comparison of hand gesture descriptions between zero-shot and \textbf{\textit{FiGOP}}-augmented models.
}
\label{case_study}
\vspace{-1.5em}
\end{figure*} 

\begin{table}[t]
\centering
\caption{Comparison of \textit{HandJudge} scores (0–5) from Qwen2.5, GPT-4.1, and human judges. }
% \vspace{-0.5em}
\label{tab:judge_agreement}
\footnotesize
\begin{tabular}{p{6em}|ccc} 
\toprule
\diagbox[
  width=7em,
  height=1em,
  font=\scriptsize,
  innerleftsep=1pt,
  innerrightsep=1pt
]{\raisebox{-1ex}{\textbf{Model}}}{\raisebox{1ex}{\textbf{Judge}}} 
    & \textbf{Qwen2.5-7B~\cite{qwen2.5}} 
    & \textbf{GPT-4.1~\cite{achiam2023gpt}} 
    & \textbf{Human} \\
\midrule
\textit{Zero-shot}  & 2.08 & 1.55 & 1.10 \\
\textit{MM Projector} & 3.30 & 3.18 & 3.01 \\
\textit{\textbf{FiGOP}} & \textbf{3.52} & \textbf{3.61} & \textbf{3.74} \\
\bottomrule
\end{tabular}
\vspace{-1.5em}
\end{table}

\subsection{Discussion}

\noindent \textbf{Evaluation Reliability.}  
We randomly sample 100 clips and collect captions from three systems: the \textit{zero-shot Qwen3-VL-8B}~\cite{qwen3technicalreport}, \textit{multimodal projector (MM Projector)}  fine-tuned model, and our \textit{FiGOP}-augmented model. 
Each caption is scored by three independent judges: \textit{Qwen2.5-7B}~\cite{qwen2.5}, \textit{GPT-4.1}~\cite{achiam2023gpt}, and human annotators, with human judgments gathered from 5 independent raters. 
The evaluations follow the four \textit{HandJudge} criteria (FHI, MT, CI, CMS; 0–5). 
As shown in Table~\ref{tab:judge_agreement}, the scores from GPT-4.1 closely match those of the human raters across all models, demonstrating the reliability of LLM-based evaluations.
Our \textit{FiGOP}-augmented model outperforms other systems and attains the highest average score.
These results indicate that the \textit{FiGOP}-augmented model yields the most structurally accurate, consistent, and interaction-aware descriptions.
Further details on the evaluation setup and protocol are provided in the \textit{Appendix}.

\noindent\textbf{Generalization under Distribution Shifts.}
To further examine model robustness, we evaluate on two out-of-distribution (OOD) subsets from \textit{BSL} (linguistic variation) and \textit{HOI4D + MotionBench} (physical variation). 
We compare the \textit{zero-shot Qwen3-VL-8B}~\cite{qwen3technicalreport}, \textit{multimodal projector (MM Projector)}  fine-tuned model, and our \textit{FiGOP}-augmented model.  
As shown in Table~\ref{tab:ood_result}, all metrics drop notably compared to the in-distribution test set, reflecting the inherent difficulty of generalizing to unseen sign languages or novel manipulation domains. 
Nevertheless, our method consistently outperforms both zero-shot and standard fine-tuned baselines across all metrics, indicating that incorporating structured motion cues improves resilience to distribution shifts. 
However, performance gaps remain large, particularly in the \textit{HandJudge} overall score, suggesting that fine-grained reasoning about unseen finger configurations and contact dynamics remains an open challenge.

\begin{table}[t]
\centering
\caption{
Comparison of model performance on out-of-distribution (OOD) subsets from BSL, HOI4D, and MotionBench. 
}
% \vspace{-0.5em}
\label{tab:ood_result}
\footnotesize
\begin{tabular}{l|cccc|c}
\toprule
\textbf{Model} & \textbf{B-4} & \textbf{R-L} & \textbf{M} & \textbf{C} & \textbf{HandJudge} \\
\midrule
\textit{Zero-shot} & 1.33&20.13&27.57&9.67& 1.56 \\
\textit{MM Projector} & 2.53&27.38&27.33&47.66 & 1.61  \\
\textbf{\textit{FiGOP}} & \textbf{3.11}&\textbf{30.47}&\textbf{29.15}&\textbf{52.21}&\textbf{2.04} \\
\bottomrule
\end{tabular}
\vspace{-1.5em}
\end{table}

\subsection{Case Study}

We compare the zero-shot model and our proposed \textit{FiGOP}-augmented version of \textit{Qwen3-VL-8B-Instruct}~\cite{qwen3technicalreport} on a hand gesture task involving the right index finger tapping the left thumb. 
The zero-shot model generates a vague description, failing to specify the fingers involved and the contact point. The GPT-4.1~\cite{achiam2023gpt} evaluation gives low scores, due to missing details such as finger identity and motion sequence.
In contrast, the \textit{FiGOP}-augmented model provides a detailed and accurate description, specifying the left hand’s position and the exact contact between the right index finger and left thumb. GPT-4.1 rates this highly, with perfect scores (5.0) in all dimensions.
These results demonstrate the significant improvement in fine-grained motion description with the \textit{FiGOP} augmentation.
% Further case studies can be found in the \textit{Appendix}.

% \section{Limitation and Future Work}

\section{Conclusion}
Understanding the subtleties of human hand motion requires a bridge between perception, language, and action.
In this work, we take a step in that direction by introducing the \textbf{\textit{FingerCap}} task, the first benchmark and framework for fine-grained finger-level motion description.
Our \textbf{\textit{FingerCap-40K}} dataset combines the linguistic precision of gesture with the physical realism of hand–object interaction, providing a foundation for studying how models interpret manual dexterity.
We further propose the \textbf{\textit{FiGOP}} module, which injects structured pose dynamics into video representations, enabling models to reason not only about visual appearance but also about temporal movement patterns at the finger level.
Through evaluations using our \textbf{\textit{HandJudge}} framework, we show that while existing Video-LLMs struggle with subtle motion understanding, incorporating explicit motion structure significantly narrows the gap.
We hope this work inspires the community to move beyond coarse global actions and toward the rich, structured language of the human hand.

\clearpage

\clearpage
\setcounter{page}{1}
% \maketitlesupplementary
\appendix

\section*{Appendix}

\noindent This Appendix is organized as follows:

\begin{itemize}
    \item Broader Impact (Section~\ref{sec:Broader_Impact})
    \item Limitations and Future Work (Section~\ref{sec:Limitations_and_Future_Work})
    \item \textit{FingerCap-40K} Data Samples (Section~\ref{sec:data_samples})
    \item Details of \textit{FiGOP} Implementation (Section~\ref{sec:implementation_details})
    \item Details of Zero-shot Generation (Section~\ref{sec:zeroshot_details})
    % \item Prompt Template for Caption Generation
    % \item Prompt Template for Caption Rephrase
    \item Details of \textit{HandJudge} Evaluation Protocol (Section~\ref{sec:handjudge_details})
    % \item Prompt Template for Evaluation
    \item Details of Human Evaluation (Section~\ref{sec:human_eval_details})
    \item Additional Ablation Studies (Section~\ref{sec:ablation_studies})
    \item Additional Case Studies (Section~\ref{sec:case_studies})
    \item Ethics Statement (Section~\ref{sec:Ethics_Statement})
\end{itemize}

\section{Broader Impact}
\label{sec:Broader_Impact}
Our work introduces \textbf{\textit{FingerCap}}, a new task for fine-grained finger-level hand motion captioning, together with the \textbf{\textit{FingerCap-40K}} dataset, the \textbf{\textit{FiGOP}} architecture, and the \textbf{\textit{HandJudge}} evaluation protocol.
\textit{FingerCap} brings several positive impacts to the broader computer vision and multimodal research community. 
By shifting human motion understanding from coarse action labels or holistic hand gestures toward the precise and interpretable dynamics of individual fingers, \textit{FingerCap} enables a more nuanced characterization of hand interactions and supports tasks that previously lacked reliable benchmarks, including detailed motion instruction and subtle behavior analysis. 
\textit{FingerCap-40K} and \textit{FiGOP} further benefit assistive technologies such as sign language understanding and fine hand operation, where accurate descriptions of finger movements are essential for conveying semantic distinctions, assessing motor progress, and guiding dexterous manipulation. 
Our findings also show that current video multimodal large language models (Video-MLLMs) struggle with fine temporal cues, which underscores the importance of explicit motion modeling in the design of future multimodal systems with potential impact on embodied intelligence, physical reasoning, and egocentric hand activity analysis. 
We hope that \textit{FingerCap} and \textit{FingerCap-40K} will inspire responsible and inclusive research toward safer and more expressive fine-grained human hands motion understanding.

\section{Limitations and Future Work} \label{sec:Limitations_and_Future_Work} Despite the promising results, our work has limitations that highlight important avenues for future research. These limitations stem from both the current dataset scope and the inherent trade-offs in our modeling framework:

\begin{itemize} 
\item \textbf{Dataset Diversity:} The current \textit{FingerCap-40K} dataset primarily covers sign language and clean, indoor hand-object interactions. 
It does not yet extensively include in-the-wild scenarios with extreme lighting, motion blur, or highly cluttered backgrounds. 
Expanding the benchmark to diverse, unconstrained environments is a crucial next step to ensure robust real-world generalization.

\item \textbf{Dependency on 2D Poses:} 
To achieve high temporal resolution with low computational cost, the \textit{FiGOP} module relies on 2D hand pose.
While efficient, this design inherits the limitations of 2D pose detectors, particularly under severe self-occlusion or depth ambiguity.
Future work could mitigate this by incorporating 3D pose priors~\cite{jiang2024rtmw} or hand mesh recovery~\cite{yin2025smplest}, provided that the computational overhead remains manageable.

\item \textbf{Data Scale and Overfitting:} Our two-stage SFT strategy significantly improves domain-specific performance but carries a risk of overfitting to the captioning style of the training set.
Exploring scale-up strategies, such as incorporating larger-scale noisy web video-text pairs with weak finger-level supervision or leveraging instruction tuning across broader motion domains, may enhance the model's generalization capabilities.

\item \textbf{Evaluation Proxy:} \textit{HandJudge} provides a necessary leap forward in evaluating finger-level semantics, yet it remains an LLM-based proxy.
Although we demonstrate high alignment with human judgment, purely text-based metrics cannot fully verify physical plausibility.
Future evaluation protocols might benefit from hybrid schemes that combine LLM scoring with physics-based verification or human-in-the-loop assessment.

\item \textbf{Scope of Generation:} 
While inverse tasks, such as generating hand motion from text, are highly relevant~\cite{hoi_gen1,hoi_gen2,sign_gen1,sign_gen3}, they lie outside the scope of this descriptive work.
We hope \textit{FingerCap-40K} will serve as a foundational resource to support future research in controllable motion generation and embodied agents.
\end{itemize}

\begin{figure*}[!h]
    \centering
    \includegraphics[width=0.95\linewidth]{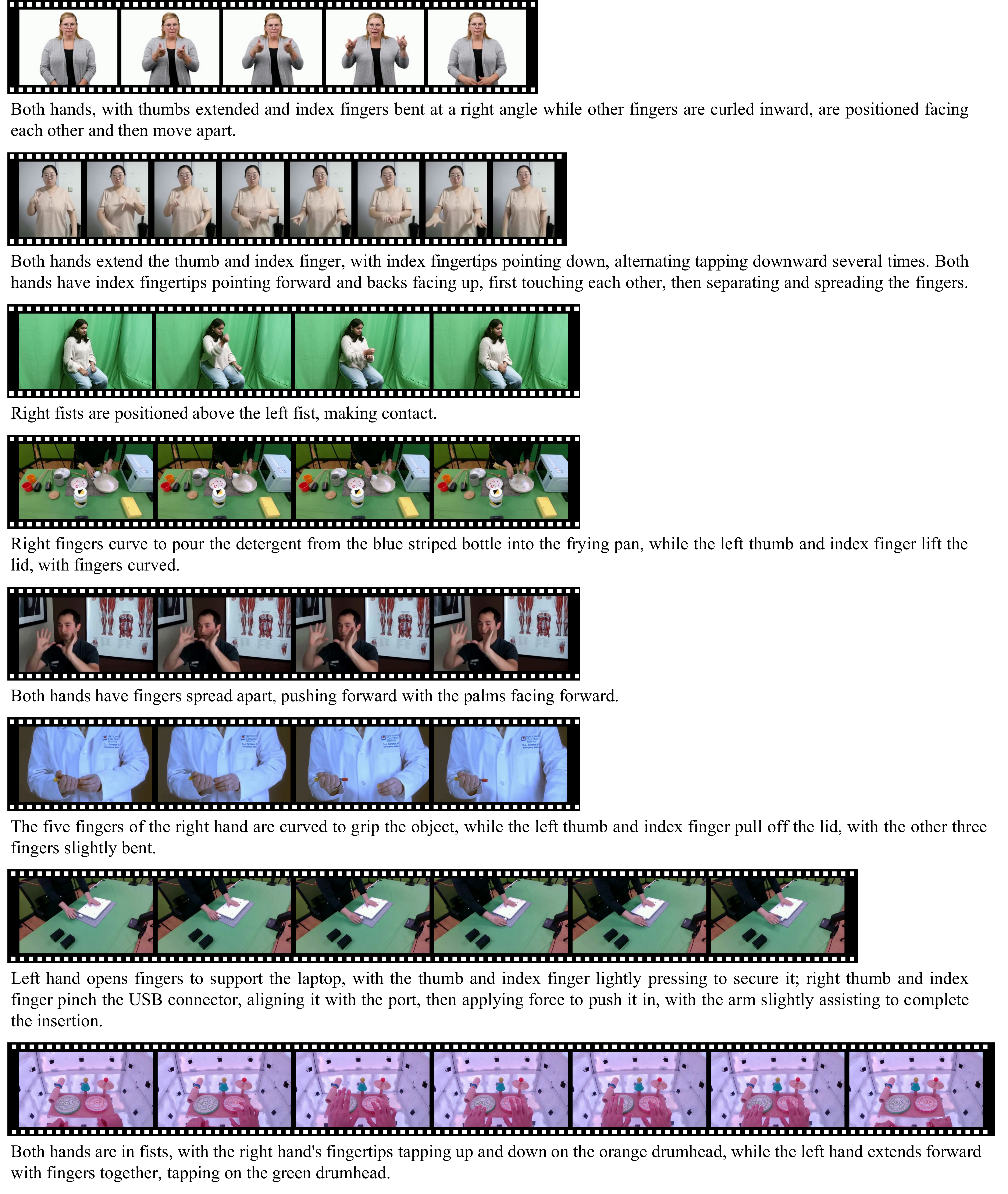} 
    \caption{
    Representative data samples from \textit{FingerCap-40K}.
    The dataset spans diverse domains including communicative gestures and physically grounded hand-object interactions.
    The corresponding captions provide dense, finger-level descriptions of joint states, spatial relationships, and precise contact dynamics.
    This diversity and granularity pose significant challenges for current Video-MLLMs and highlight the unique value of our benchmark.
    }
    \label{fig:more_samples}
\end{figure*}

\section{\textit{FingerCap-40K} Data Samples}
\label{sec:data_samples}

We provide additional data examples from the \textit{FingerCap-40K} dataset to demonstrate the diversity of data sources, the complexity of hand interactions, and the granularity of our textual annotations.

\noindent\textbf{Diversity in Scenarios and Demographics.}
As shown in Figure~\ref{fig:more_samples}, our dataset covers a wide range of visual domains. 
The samples include controlled environments with green screens, clean indoor studios, and cluttered egocentric or third-person workspaces.
Consistent with our \textit{Ethics Statement}, the samples also reflect a diverse distribution of participants across different genders and skin tones, ensuring the model learns robust representations that generalize across demographic groups.

\noindent\textbf{Fine-grained Finger-level Semantics.}
Unlike traditional human motion captioning datasets that focus on high-level action, \textit{FingerCap-40K} provides explicit descriptions of finger articulation.
For instance, the fourth row details how the \textit{``right fingers curve to pour''} while the \textit{``left thumb and index finger lift the lid''}, capturing the bimanual coordination required for the manipulation task.
Similarly, gesture examples describe precise joint configurations, which are essential for distinguishing subtle lexical meanings.

\noindent\textbf{Complex Object Interactions.}
Our dataset also includes challenging manipulation tasks involving small objects and fine motor control. 
Row 7 demonstrates a multi-stage interaction: \textit{``pinching the USB connector''}, \textit{``aligning it''}, and \textit{``pushing it in''}, which requires the model to understand contact, force, and trajectory at a microscopic level.

\section{Details of \textit{FiGOP} Implementation}
\label{sec:implementation_details}

We elaborate on the architecture and data flow of the \textit{FiGOP}-augmented Video-MLLM. 
We focus on the interaction between the visual and pose streams during the encoding and fusion stages.

\subsection{Model Configuration and Input}
Our model processes video inputs using a slow-fast sampling strategy.
For a standard inference setting (Batch Size = 1), the input consists of:
\begin{itemize}
    \item \textbf{Slow Visual Stream:} We sample $T=15$ RGB keyframes from the video. Each frame is resized to $448 \times 448$.
    \item \textbf{Fast Pose Stream:} Associated with each RGB keyframe is a high-frequency pose clip of length $T_p=8$. The pose representation covers $J=42$ hand joints (bimanual) with 3 channels ($x, y, \text{confidence}$).
\end{itemize}

\subsection{Architecture of Data Flow}
The information processing pipeline consists of three parallel stages: visual encoding, pose motion encoding, and cross-modal fusion.

\noindent\textbf{Visual Encoding.}
The RGB keyframes are processed by the frozen vision tower (Qwen3-VL Vision Transformer~\cite{dosovitskiy2020image,qwen3technicalreport,Qwen-VL}). 
The images are tokenized and encoded into visual feature sequence $F^v \in \mathbb{R}^{T \times N \times D_{llm}}$, where $N=256$ is the number of visual tokens per frame and $D_{llm}=4096$ is the hidden dimension of the LLM~\cite{qwen3technicalreport}.

\noindent\textbf{Pose Motion Encoding.}
The pose stream aims to capture fine-grained motion dynamics that are lost in the sparse visual stream.
The input pose tensor $P \in \mathbb{R}^{T \times T_p \times J \times 3}$ is processed as follows:
\begin{itemize}
    \item \textbf{Spatial Modeling:} A multi-layer ST-GCN~\cite{shen2025cross,cosign,stgcn} first aggregates spatial information across the $J$ joints for each frame, projecting the joint topology into a latent feature space $D_{pose}=256$.
    \item \textbf{Temporal Modeling:} The spatially aggregated features are then fed into a lightweight Temporal Transformer~\cite{vaswani2017attention}. This module models the evolution of hand states within the $T_p$-frame window, producing the motion embeddings $F^p \in \mathbb{R}^{T \times T_p \times D_{llm}}$ (after projection).
\end{itemize}
Crucially, this stream operates independently on each \textit{FiGOP} unit, preserving the local temporal correspondence with the visual keyframes.

\noindent\textbf{Motion-Aware Fusion.}
To inject fine-grained dynamics into the visual representation, we employ a Motion-Aware Projector.
We utilize a Cross-Attention~\cite{vaswani2017attention} mechanism:
\begin{itemize}
    \item \textbf{Query ($\mathbf{Q}$):} Derived from the visual features $F^v$.
    \item \textbf{Key/Value ($\mathbf{K}, \mathbf{V}$):} Derived from the pose motion embeddings $F^p$.
\end{itemize}
This design allows each static visual token to attend to the dense motion history surrounding it. 
The fused features are then projected to the LLM's input space, flattened, and concatenated with text embeddings for autoregressive generation~\cite{qwen3technicalreport,Qwen-VL}.

\subsection{Parameter Efficiency}
A core advantage of our design is its efficiency. 
Instead of using heavy 3D video backbones~\cite{wang2022internvideo,wang2024internvideo2}, \textit{FiGOP} leverages lightweight pose encoders. 
The total number of trainable parameters (including the pose encoder and the motion-aware projector) is approximately \textbf{248M}. 
Compared to the \textbf{8B} parameters of the frozen backbone, this represents a marginal increase of only $\sim \mathbf{3\%}$, enabling fine-grained motion understanding with minimal computational overhead.

\section{Details of Zero-shot Generation}
\label{sec:zeroshot_details}

To ensure a rigorous and fair comparison across diverse open-source and proprietary Video-MLLMs (including GPT-4o~\cite{hurst2024gpt}, Gemini-2.5-Pro~\cite{comanici2025gemini}, and the Qwen~\cite{qwen2.5-VL}/InternVL~\cite{internvl3_5} families), we establish a unified generation protocol. 
This protocol standardizes the decoding parameters and the instruction pipeline to decouple the models' motion understanding capabilities from their conversational styles.

\subsection{Decoding Configuration}
For all zero-shot evaluations, we adopt a consistent, low-temperature decoding strategy to balance descriptive diversity with factual determinism. Specifically, we set the \textit{temperature} to 0.2 and \textit{top-p} to 0.9.
This setting encourages the models to ground their generation strictly in visual evidence, minimizing hallucinations common in high-temperature sampling.

\subsection{Prompting Strategy}
\textbf{Generation Phase.} 
We utilize a structured system prompt, shown in Figure~\ref{fig:prompt_template}, to instruct the models. 
Crucially, this prompt enforces a strict coordinate system definition: ``left'' and ``right'' must always refer to the actor's body orientation rather than the camera view. 
This is essential for consistent evaluation across first-person (egocentric) and third-person viewpoints. 
The prompt also explicitly requests details on finger states and contact dynamics.

\noindent\textbf{Rephrasing Phase for Fairness.}
We observed that zero-shot outputs from general-purpose Video-MLLMs often exhibit significant stylistic variance. 
Some models output meta-commentary (\eg, \textit{``The video displays...''}) or irrelevant visual details (\eg, \textit{``A person in a black shirt...''}), which negatively impact standard captioning metrics (BLEU~\cite{bleu}, CIDEr~\cite{vedantam2015cider}) even when the motion semantics are correct.
To mitigate this, we employ a deterministic post-processing step using GPT-4.1~\cite{achiam2023gpt}. 
As detailed in Figure~\ref{fig:prompt_rephrase}, the rephrasing prompt directs the assistant to filter out non-motion content and convert the description into a neutral, third-person tense. 
We strictly enforce a \textbf{``no new information''} rule to ensure that the rephrasing process functions solely as a stylistic normalizer and does not hallucinate new motion details.

\section{Details of \textit{HandJudge} Evaluation Protocol}
\label{sec:handjudge_details}

Conventional n-gram metrics (\eg, BLEU~\cite{bleu}, METEOR~\cite{banerjee2005meteor}) are insufficient for evaluating fine-grained motion understanding. 
For instance, misidentifying the ``index finger'' as the ``ring finger'' results in a negligible penalty in text overlap metrics but represents a catastrophic failure in physical reasoning.
To address this, we design \textbf{\textit{HandJudge}}~\cite{Gu2024AASO,li2025generation}, a reference-based evaluation protocol powered by GPT-4.1~\cite{achiam2023gpt}.

\subsection{Evaluation Dimensions}
As illustrated in the system prompt in Figure~\ref{fig:prompt_evaluation}, \textit{HandJudge} assesses model predictions against ground-truth references on a strict 0-5 scale across four dimensions:

\begin{enumerate}
    \item \textbf{Finger and Hand Identification (FHI):} Measures the anatomical precision. It penalizes ambiguity (\eg, generic ``fingers'') and explicitly checks for correct laterality (left vs. right hand) and specific joint usage.
    \item \textbf{Motion and Trajectory (MT):} Evaluates the kinematic fidelity. It distinguishes between subtle variations in movement types (\eg, ``tapping'' vs. ``pressing'') and validates directional correctness.
    \item \textbf{Contact and Interaction (CI):} Focuses on physical grounding. It verifies whether the model correctly describes the surfaces of contact (\eg, ``fingertip'' vs. ``palm'') and the interaction with objects.
    \item \textbf{Completeness of Motion Sequence (CMS):} Assesses temporal coverage. It ensures the generated caption captures the full temporal evolution (start, transition, end) rather than describing a static pose.
\end{enumerate}

\subsection{Scoring Mechanism}
To ensure interpretability, the LLM is instructed to output a rationale (\texttt{"explanation"}) before assigning numerical scores. 
This Chain-of-Thought (CoT)~\cite{Gu2024AASO,li2025generation} process encourages the judge to analyze specific discrepancies before quantifying the error, leading to higher alignment with human judgment as demonstrated in the main paper.

\section{Details of Human Evaluation}
\label{sec:human_eval_details}

To validate the reliability of our automated \textit{HandJudge} metric, we conducted a rigorous human evaluation study. 
The goal is to determine whether the LLM-based scoring aligns with human perception regarding the subtlety and precision of finger movements.

\subsection{Experimental Setup}
\textbf{Data Sampling.} 
We randomly sampled 100 video clips from the \textit{FingerCap-40K} test set, ensuring a balanced representation of both gesture and HOI scenarios.

\noindent\textbf{Model Candidates.} 
For each clip, we collected captions generated by three distinct systems representing different performance tiers:
\begin{itemize}
    \item Zero-shot Baseline: Qwen3-VL-8B-Instruct (without fine-tuning).
    \item Standard Fine-tuning: Qwen3-VL-8B + MM Projector (SFT).
    \item Ours: Qwen3-VL-8B + \textit{FiGOP} (SFT).
\end{itemize}

\noindent\textbf{Human Raters and Protocol.} We recruited 5 independent evaluators. 
To ensure consistency, all raters were trained on the \textit{HandJudge} rubric (detailed in Figure~\ref{fig:prompt_evaluation}) prior to the study. 
They were strictly instructed to grade the generated captions against the ground truth videos on the same 4 dimensions (FHI, MT, CI, CMS) using the 0-5 scale. 
In total, this resulted in $100 \text{ clips} \times 3 \text{ models} \times 5 \text{ raters} = 1,500$ individual annotations.

\subsection{Alignment Analysis}
We compared the average scores assigned by the human panel against those assigned by GPT-4.1~\cite{achiam2023gpt} (the engine behind \textit{HandJudge}). 
As reported in the main paper (Table 4), the alignment is highly consistent:

\begin{itemize}
    \item \textbf{Rank Consistency:} Both human raters and \textit{HandJudge} produced the exact same performance ranking: \textit{FiGOP  $>$  MM Projector $>$ Zero-shot}. This confirms that \textit{HandJudge} correctly discriminates between model capabilities.
    
    \item \textbf{Score Proximity:} The absolute score differences between Human and AI judges were minimal, particularly for the high-performing models. 
    For our \textit{FiGOP} model, the human score is \textbf{3.74} and the \textit{HandJudge} score is \textbf{3.61}, demonstrating a deviation of less than $4\%$. 
    
    \item \textbf{Sensitivity to Errors:} Interestingly, humans were slightly harsher on the Zero-shot baseline (1.10) compared to \textit{HandJudge} (1.55). 
    Qualitative feedback from raters suggests that humans penalize ``hallucinated'' finger details more severely than the LLM. 
    However, this implies that \textit{HandJudge} is a \textit{conservative} metric, meaning that if a model scores high on \textit{HandJudge}, it is highly likely to be perceptually accurate to humans.
\end{itemize}

In conclusion, \textit{HandJudge} serves as a scalable, reliable, and cost-effective proxy for fine-grained human hands motion evaluation, exhibiting strong correlation with expert human judgment.

\section{Additional Ablation Studies}
\label{sec:ablation_studies}

In this section, we provide further empirical analysis to justify our design choices regarding the temporal resolution of the pose stream and the composition of the training data, as well as to verify the generalization capability of our method across different Video-MLLMs.

\subsection{Impact of Pose Sequence Length}
A key hyperparameter in our \textit{FiGOP} module is the length of the dense pose sequence ($T_p$) attached to each sparse RGB keyframe.
In the main paper, we set $T_p=8$ (representing 8 pose frames per RGB keyframe).
To validate this choice, we conducted an ablation study with $T_p \in \{4, 8, 16\}$.

As shown in Table~\ref{tab:ablation_frames}, decreasing the sequence length to $T_p=4$ results in a performance drop across all metrics.
This suggests that excessively short pose windows fail to capture sufficient temporal context for complex finger articulations.
Conversely, increasing the length to $T_p=16$ does not yield further improvements and slightly degrades performance.
We hypothesize that overly long pose sequences may introduce temporal redundancy and increase the susceptibility to accumulated noise from 2D pose estimation errors (\eg, jitter or occlusion).
Such noise can distract the lightweight encoder, making it harder to focus on the most relevant high-frequency dynamics near the keyframe.
Thus, $T_p=8$ offers the optimal balance between motion granularity and modeling efficiency.

\begin{table}[h]
    \centering
    \footnotesize
    \caption{Ablation on Pose Sequence Length ($T_p$).
    % Results are reported on the full \textit{FingerCap-40K} test set (Overall). 
    % The default setting ($T_p=8$) achieves the best trade-off.
    }
    \begin{tabular}{l|cccc}
    \toprule
    \textbf{Pose Frames ($T_p$)} & \textbf{B-4} & \textbf{R-L} & \textbf{METEOR} & \textbf{CIDEr} \\
    \midrule
    4 Frames  & 14.12 & 36.10 & 37.95 & 147.50 \\
    \textbf{8 Frames} & \textbf{15.36} & \textbf{37.92} & \textbf{38.89} & \textbf{155.27} \\
    16 Frames & 15.15 & 37.75 & 38.50 & 153.80 \\
    \bottomrule
    \end{tabular}
    \label{tab:ablation_frames}
\end{table}

\subsection{Effect of Dataset Composition}
The \textit{FingerCap-40K} dataset comprises two distinct domains: \textit{Gesture} (linguistically structured) and \textit{HOI} (physically grounded).
To verify the necessity of joint training, we trained separate models on each subset and compared them with our unified model.

The results in Table~\ref{tab:ablation_domain} reveal a clear trade-off:
\begin{itemize}
    \item \textbf{Specialized Training:} Models trained exclusively on a single domain (\eg, Gesture-only) achieve slightly higher performance on their corresponding test set compared to the unified model. This is expected as the model overfits to the specific domain distribution.
    \item \textbf{Cross-Domain Failure:} However, these specialized models fail catastrophically when evaluated on the unseen domain. For instance, the Gesture-trained model drops significantly on HOI evaluation, indicating a lack of physical reasoning capabilities.
    \item \textbf{Unified Generalization:} Our joint training strategy (Mixed) maintains competitive high performance across both domains. 
While it sacrifices a marginal amount of domain-specific accuracy, it gains robust generalization capabilities, making it the superior choice for a general-purpose finger motion understanding system.
\end{itemize}

\begin{table}[t]
    \centering
    \footnotesize
    \caption{Ablation on Dataset Composition. 
    % We compare models trained on individual subsets versus the combined dataset. 
    % While domain-specific training maximizes in-domain scores, joint training enables robust generalization across both Gesture and HOI tasks.
    }
    \setlength{\tabcolsep}{4pt}
    \begin{tabular}{l|cc|cc}
    \toprule
    \multirow{2}{*}{\textbf{Training Data}} & \multicolumn{2}{c|}{\textbf{Test on Gesture}} & \multicolumn{2}{c}{\textbf{Test on HOI}} \\
    & B-4 & CIDEr & B-4 & CIDEr \\
    \midrule
    Gesture Only & \textbf{15.10} & \textbf{154.50} & 6.21 & 58.49 \\
    HOI Only     & 4.15 & 42.10 & \textbf{18.45} & \textbf{169.10} \\
    \midrule
    \textbf{Mixed} & 13.81 & 146.29 & 17.09 & 165.31 \\
    \bottomrule
    \end{tabular}
    \label{tab:ablation_domain}
\vspace{-1em}
\end{table}

\begin{table}[b]
    \centering
    \footnotesize
    \vspace{-1em}
    \caption{Generalization Analysis on Qwen2.5-VL-7B-Instruct.
    % We report the Average scores across standard captioning metrics. 
    % The consistent gains confirm that FiGOP is effective across different Video-MLLM architectures.
    }
    \begin{tabular}{l|cccc}
    \toprule
     & \textbf{B-4} & \textbf{R-L} &\textbf{METEOR} & \textbf{CIDEr} \\
    \midrule
    Zero-shot  & 1.87 & 21.81 & 26.92 & 29.06 \\
    MM Projector & 8.85& 31.45 & 32.10 & 101.45 \\
    \textbf{\textit{FiGOP} (Ours)} & \textbf{13.52} &\textbf{35.80} & \textbf{36.65} &  \textbf{148.20} \\
    \bottomrule
    \end{tabular}
    \label{tab:qwen2.5_ablation}

\end{table}

\subsection{Generalization to Other Video-MLLMs}

A core advantage of the \textit{FiGOP} architecture is its model-agnostic design. 
While our main experiments utilize Qwen3-VL-8B as the backbone, the lightweight pose stream and motion-aware projector can be seamlessly integrated into other Video-MLLMs.
To verify this, we applied our method to \textbf{Qwen2.5-VL-7B-Instruct}~\cite{qwen2.5-VL}.
We compare three settings on the full \textit{FingerCap-40K} test set: 
(1) Zero-shot baseline; 
(2) Standard Supervised Fine-tuning (SFT) with a vanilla Multimodal Projector; 
and (3) Our \textit{FiGOP}-augmented SFT.

As presented in Table~\ref{tab:qwen2.5_ablation}, the results mirror the trends observed with the Qwen3 backbone.
Specifically, standard fine-tuning significantly boosts performance over the zero-shot baseline, demonstrating the necessity of domain adaptation. 
More importantly, incorporating \textit{FiGOP} yields a further substantial improvement, increasing the CIDEr score to \textbf{148.20}.
These findings confirm that \textit{FiGOP} provides a consistent benefit for fine-grained motion understanding, independent of the specific underlying LLM architecture.

\section{Additional Case Studies} 
\label{sec:case_studies}

In this section, we provide a qualitative comparison between our \textbf{\textit{FiGOP}-augmented model}, the \textbf{Zero-shot baseline} (Qwen3-VL-8B-Instruct~\cite{qwen3technicalreport,qwen2.5-VL}), and a state-of-the-art proprietary model (\textbf{Gemini-2.5-Pro}~\cite{comanici2025gemini}).
As shown in Figure~\ref{fig:qualitative_cases}, we select three representative scenarios covering sign language gestures, dynamic object manipulation, and tool use.

\noindent\textbf{Case 1: Complex Finger Configuration (Top Row).}
The first example features a precise sign language gesture involving an asymmetric hand shape. 
The \textbf{Gemini-2.5-Pro} model generates a fluent but hallucinated description, incorrectly stating that the hands ``clasp'' and ``unclasp'', missing the critical semantic detail of the right index finger pointing to the left pinkie.
The \textbf{Zero-shot baseline} fails to identify the specific finger contact targets.
In contrast, our \textbf{\textit{FiGOP}-augmented model} accurately captures the fine-grained static pose: \textit{``right fist is placed directly on top of the left fist''} and explicitly identifies the \textit{``index finger extended, pointing to the pinkie finger''}, demonstrating superior geometric reasoning.

\noindent\textbf{Case 2: Rapid Dynamic Interaction (Middle Row).}
This case involves a magic trick with a coin (``flicking a fake coin''), characterized by rapid, high-frequency motion.
The \textbf{Zero-shot baseline} suffers from severe repetition loops (repeating ``moves the coin'' multiple times), a common failure mode when sparse visual tokens fail to resolve fast temporal changes.
\textbf{Gemini-2.5-Pro} hallucinates a ``deck of cards'' which is not part of the active interaction.
Our model, leveraging the dense pose stream, efficiently summarizes the dynamic action: \textit{``pinches the edge of a coin''} and \textit{``flicks the fake coin into the air''}, effectively capturing the causality of the motion.

\noindent\textbf{Case 3: Action-Object Disambiguation (Bottom Row).}
The final example shows a user sharpening a pencil. This is a challenging case where the object (a small block sharpener) is occluded and ambiguous.
The \textbf{Zero-shot baseline} misidentifies the action as ``pressing a stamp'', likely relying on static visual appearance.
\textbf{Gemini-2.5-Pro} describes the motion vaguely as moving a ``small brown object'' in a ``circular motion''.
However, our \textbf{\textit{FiGOP}} model correctly grounds the fine-grained motion cues (twisting/inserting) to disambiguate the object, correctly identifying both the \textit{``pencil''} and the \textit{``sharpener''}, and describing the action of \textit{``placing the pencil inside''}.

\section{Ethics Statement}
\label{sec:Ethics_Statement}

This work involves the curation of \textbf{\textit{FingerCap-40K}}, a large-scale dataset designed for fine-grained finger-level hand motion captioning.
We have carefully considered the ethical implications of data collection, annotation, and release, ensuring strict compliance with the \textit{CVPR Ethics Guidelines}.

\textbf{Human Subjects \& Consent.}
Our dataset is derived exclusively from established, publicly available datasets and text corpora that are explicitly licensed for academic research.
We have verified that the original source datasets (including ASL, BSL, CSL, Auslan, GigaHands, and OakInk2) obtained necessary consents from participants during their initial collection.
% No new video recordings of human subjects were conducted specifically for this project; our contribution lies in the fine-grained annotation and linguistic refinement of these existing resources.

\textbf{Compensation.}
All annotators involved in the data cleaning, description verification, and refinement process were fairly compensated.
Contributors were paid at a rate of \textbf{\$50 USD per hour}, which exceeds the local minimum wage and aligns with institutional fair labor standards.
In total, approximately 300 hours of paid annotation work were carried out under formal contracts to ensure high-quality, expert-verified data.

\textbf{Privacy \& Anonymization.}
We strictly adhere to the privacy protocols of the original data sources.
All participants in the aggregated datasets have previously consented to the public release of their recordings for academic use.
Nevertheless, we have implemented a robust withdrawal and anonymization protocol: we apply face-blurring (using \href{}{\texttt{deface}}) upon any participant's request and will remove data entirely if consent is withdrawn.
Future releases will also prioritize 2D/3D pose annotations to support privacy-preserving research.
Crucially, no personally identifiable information (PII) beyond the visual data itself, nor any sensitive data (such as health or financial information), is collected or maintained.

\textbf{Copyright \& Licensing.}
All curated video segments are distributed under the \textbf{CC BY-NC-SA 4.0} license, consistent with the original data sources.
The dataset is released with an accompanying End-User License Agreement (EULA) that explicitly prohibits commercial exploitation, unauthorized re-identification, and usage in surveillance systems.
A dedicated contact email is provided in the repository for takedown or anonymization requests.

\textbf{Fairness \& Representativeness.}
As illustrated in Figure~2 (main paper) and Figure~\ref{fig:more_samples}, we explicitly addressed demographic bias during data curation.
The dataset encompasses a diverse range of human subjects, covering various skin tones, genders, and hand shapes, to ensure the model's robustness and fairness across different demographic groups.
By integrating sign languages from multiple regions (ASL, BSL, CSL, Auslan), we also aim to capture cultural and linguistic diversity in hand motion.

\textbf{Responsible Use.}
We include a Responsible Use Statement with the dataset that explicitly prohibits its deployment in surveillance, biometric identification, or other sensitive decision-making contexts without further ethical review.
Our release aims to support inclusive, equitable research benefiting the multimodal understanding and assistive technology communities.

\begin{figure*}[t] % [t] 表示优先放在页面顶部，* 表示跨双栏
\centering
\begin{promptbox}[Prompt Template for Caption Generation]{lightblue}{prompt:caption-generation}

\textbf{System:}

\begin{quote}
    You are an expert in describing fine-grained hand and finger motions in videos.
\end{quote}

\textbf{User:} 
\begin{quote}
Your task is to provide detailed, specific descriptions of **hand and finger motions**—including actions, trajectories, and interactions between hands or with objects. \\

Videos may use first-person (head-mounted camera), third-person (external view), or other angles. Critically, "left" and "right" must always refer to the **person's own body orientation** in the frame: never the camera's or viewer's perspective. In first-person views, left/right aligns with the person's actual left/right hands. In third-person or angled shots, judge based on the person's body direction, not the visual left/right of the frame.\\

\textbf{Requirements}:
\begin{itemize}
    \item Focus solely on visible hand/finger **motion** and **contact** with the other hand or object.
    \item Specify which hand (left, right, or both) performs the action.
    \item Detail each finger's state and movement (e.g., extended, bent, curled, touching, pinching), naming the finger explicitly (e.g., thumb, index finger).
    \item Precisely describe interactions with objects or the other hand (e.g., holding, tapping, sliding, grasping, pushing, pulling, etc.).
    \item Exclude all references to the person's face, body, background, clothing, camera, or context outside hands/fingers.
    \item Use concise English sentences; avoid bullet points, lists, captions, or summaries.
    \item Describe actions directly without speculation or analysis.
    \item Elaborate on the dynamic process of movements.\\
\end{itemize}

Here are some examples of the description:
\begin{itemize}
    \item Left thumb and index finger pinch one corner of the tablecloth, while the middle, ring, and little fingers lightly touch the surface, with the forearm slightly raised, both hands working together to lift the tablecloth.
    \item Left index and middle fingers rest on the piano; right thumb and index finger pluck the piano strings, gradually pressing down with the arm and slowing the alternation.
    \item Right hand grips a stack of cards while the left hand rubs the cards, fanning them out.
    \item Both hands open with fingers spread, the right middle finger moves slightly forward while the left fingers remain naturally spread.
    \item Both hands are in fists with pinkies extended, and the right hand moves downward to touch the pinky of the left hand.
    \item The right index finger crosses over the left index finger, and both fingers are placed on the chest, then both index fingers move downward twice.
    \item Both hands have all five fingers extended with palms facing each other, and the right hand uses the middle finger to touch the palm of the left hand twice.\\
\end{itemize}

Now, please refer to the examples above and describe the movements of the hands and fingers in this video according to the given requirements.

\end{quote}
\end{promptbox}
\caption{The system prompt used for zero-shot generation.}
\label{fig:prompt_template} % 这是你的 Label
\end{figure*}

\begin{figure*}[t] % [t] 表示优先放在页面顶部，* 表示跨双栏
\centering
\begin{promptbox}[Prompt Template for Caption Rephrase]{lightblue}{prompt:caption-rephrase}

\textbf{System:}

\begin{quote}
    You are an assistant specializing in rewriting hand and finger motion descriptions to be neutral and precise.
\end{quote}

\textbf{User:} 
\begin{quote}
\textbf{Goals}

Your task is to refine sentences containing redundant descriptive elements (e.g., ``in this video'', ``this woman'') into accurate, objective action accounts, in strict compliance with the following logical rules.
\textbf{Rules}
\begin{itemize}
    \item Describe only **actual hand and finger movements**—exclude symbolic meanings, emotions, cultural connotations, or any extraneous details related to the video itself or the characters.
    \item Use **third-person descriptive tense** for all motions; avoid imperative language.
    \item Ensure all content is directly paraphrased from the original text; do not add new information.
    \item Employ concise, fluent English while retaining all relevant hand and finger details—including specific fingers involved and movement specifics.
    \item Output only the final English sentence as a single textbf.
    \item Omit labels, numbering, explanations, or any supplementary content.
\end{itemize}

\textbf{Example 1}
\begin{itemize}
    \item Original description: In the video, the person's left forearm is positioned in front of their body with the palm facing inward. Meanwhile, the right hand, with all five fingers fully extended and the palm facing forward, moves behind the left hand and begins shaking back and forth, waving the palm outward repeatedly throughout the sequence. The left forearm maintains its initial position without significant movement while the right hand continues its back-and-forth waving motion. There is no interaction with any objects or additional hand movements visible in the frame.
    \item Rewritten description: Left forearm is positioned in front of the body with the palm facing inward, while the right hand, with all five fingers extended, moves behind the left hand with the palm facing forward, shaking back and forth with the palm waving outward.
\end{itemize}

\textbf{Example 2}
\begin{itemize}
    \item Original description: In the video, both of the person's hands are featured at chest level in front of their body. Each hand extends with the index finger pointing forward and the remaining fingers closed tightly. As the sequence progresses, the index fingers of both hands move toward each other alternately, creating a motion that appears as if they are striking one another. The closed fingers on both hands remain in their initial state without any additional movement, and there is no interaction with external objects during these motions.
    \item Rewritten description: Both hands extend with index fingers pointing forward and other fingers closed, held in front of the body at chest level, while the index fingers move toward each other alternately, as if striking each other.
\end{itemize}

\textbf{Example 3}
\begin{itemize}
    \item Original description: In the video, the person's left hand is shown over a table. The left thumb and index finger are used to pinch one corner of the tablecloth. Meanwhile, the middle, ring, and little fingers lightly touch the table surface. The forearm is slightly raised. The clip shows both hands working together, and the main action involves them lifting the tablecloth. The hands maintain this position as they lift.
    \item Left thumb and index finger pinch one corner of the tablecloth, while the middle, ring, and little fingers lightly touch the surface, with the forearm slightly raised, both hands working together to lift the tablecloth.
\end{itemize}

\textbf{Output Format (STRICT)}
\begin{itemize}
    \item Your entire response must be a single, valid JSON object.
    \item Do not include any introductory phrases or any explanations.
    \item The final format should look like this:
    \begin{verbatim}
```json
{
  "rewritten_description": "Rewritten description here."
}
```
    \end{verbatim}\\
\end{itemize}

Now, please refer to the examples above and rewrite the description to be neutral and precise.

Original description: \texttt{\{original\_description\}}
\end{quote}
\end{promptbox}
\caption{The system prompt employed for the caption rephrasing stage.}
\label{fig:prompt_rephrase} % 这是你的 Label
\end{figure*}

\begin{figure*}[t] % [t] 表示优先放在页面顶部，* 表示跨双栏
\centering
\begin{promptbox}[Prompt Template for Evaluation]{lightblue}{prompt:caption-evaluation}
\label{template:caption-evaluation}

\textbf{System:}

\begin{quote}
    You are a highly critical evaluator specializing in sign language and fine-grained hand-object motion analysis.
\end{quote}

\textbf{User:} 
\begin{quote}
Your task is to rate the description provided by the model based on the given reference action description. 

The specific requirements are as follows:

\textbf{Evaluation Dimensions (STRICT, 0-5 scale)}
\begin{itemize}
    \item Finger and Hand Identification (0-5): Evaluate how precisely the description provided by the model identifies **which hand(s)** and **which fingers** are involved, including their physical state.
        \begin{itemize}
            \item 5 = Both the hand side and specific fingers are correctly identified; finger states are consistent with the reference. 
            \item 4 = Correct hand side(s) and approximate finger count, but missing exact finger naming or minor state omissions. 
            \item 3 = Only general mention of fingers or hands without specificity.
            \item 2 = Partial or ambiguous hand assignment, or wrong number of fingers.  
            \item 1 = Incorrect hand side or clearly mismatched fingers. 
            \item 0 = No mention of which hand or fingers at all.  
        \end{itemize}
    \item Motion and Trajectory (0-5): Evaluate correctness of motion type, direction, and order.
        \begin{itemize}
            \item 5 = Exact match in motion type, direction, and sequence.
            \item 4 = Slight deviation (e.g., “upward” vs “slightly forward”).
            \item 3 = Correct general motion type, but direction/order off.
            \item 1-2 = Wrong trajectory type or reversed sequence.
            \item 0 = Motion unrelated or missing.
        \end{itemize}
    \item Contact and Interaction (0-5): Evaluate hand-to-hand, hand-to-body, or hand-to-object contact.
        \begin{itemize}
            \item 5 = Fully accurate contact: correct surfaces, sides, and contact type (tap, rub, grasp, hold, etc.).
            \item 4 = Minor omission (e.g., misses which side of palm).
            \item 3 = Vague mention of contact but missing type or surface.
            \item 1-2 = Contact described incorrectly (wrong body part/object).
            \item 0 = No interaction or incorrect contact entirely.
        \end{itemize}
    \item Completeness of Motion Sequence (0-5): Evaluate whether all motion stages (start → transition → end) are covered.
        \begin{itemize}
            \item 5 = Fully complete; includes start, main action, and resolution.
            \item 4 = Minor transitional details omitted.
            \item 3 = Partial sequence (missing either start or end).
            \item 1-2 = Major gaps or only one static posture mentioned.
            \item 0 = Very incomplete; not a valid motion sequence.
        \end{itemize}
\end{itemize}

\textbf{Output Format (STRICT)}
\begin{itemize}
    \item Your entire response must be a single, valid JSON object.
    \item Do not include any introductory phrases or any explanations.
    \item The final format should look like this:
    \begin{verbatim}
{
  "explanation": string,
  "finger": float,
  "motion": float,
  "contact": float,
  "completeness": float
}
    \end{verbatim}\\
\end{itemize}
The following is the reference description:

\texttt{\{reference\}}

The following is the description provided by the model:

\texttt{\{prediction\}}

Now, please rate the description provided by the model according to the above requirements.

\end{quote}
\end{promptbox}
\caption{The system prompt employed for the \textit{HandJudge}.}
\label{fig:prompt_evaluation} % 这是你的 Label
\end{figure*}

\begin{figure*}[t]
    \centering
    \includegraphics[width=0.9\linewidth]{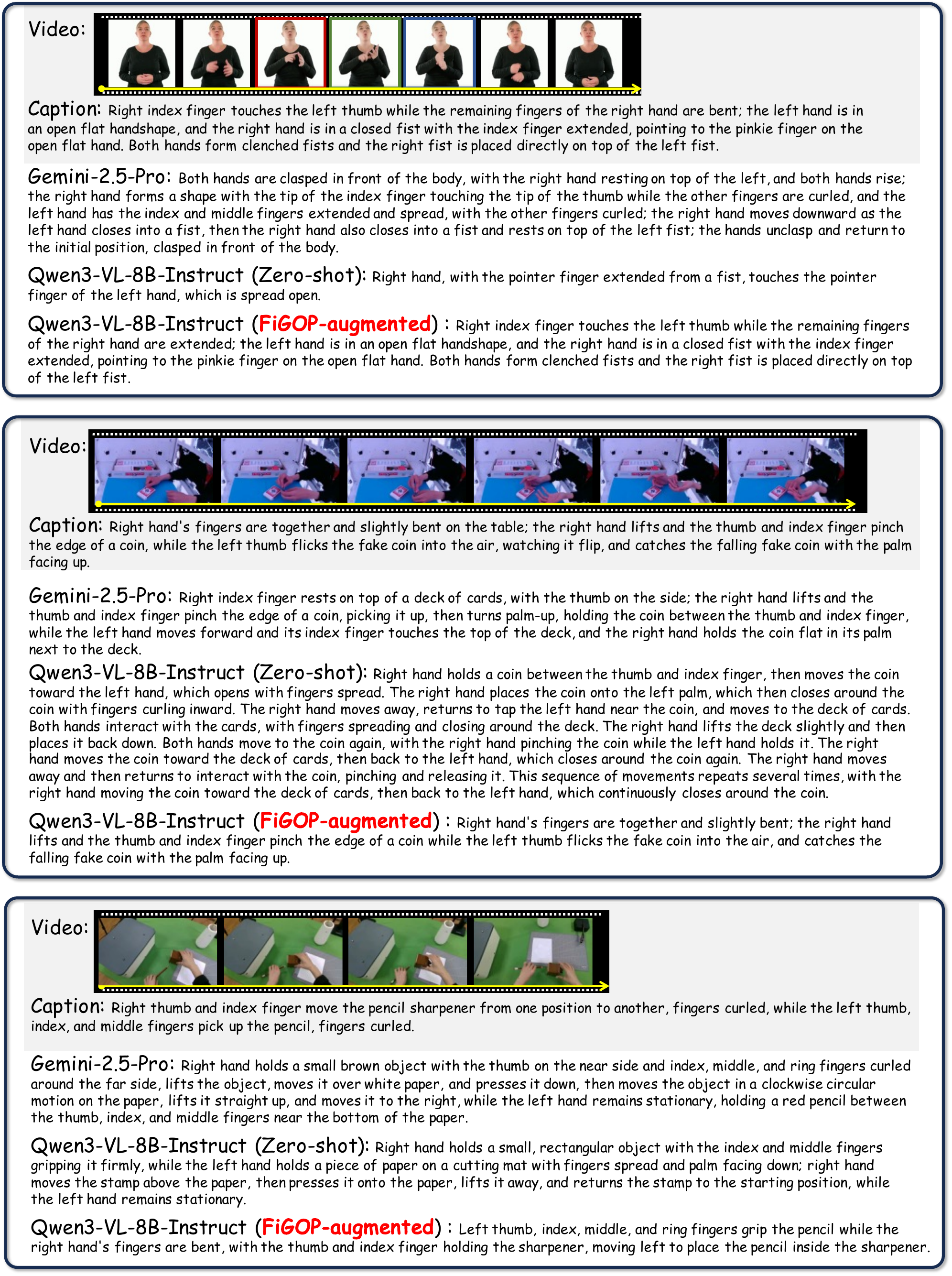} 
    \caption{
    Additional case studies on \textit{FingerCap-40K} test set. 
    % We compare the generation results of Gemini-2.5-Pro, Qwen3-VL-8B-Instruct (Zero-shot), and our \textit{FiGOP}-augmented model. 
    % The examples highlight our method's ability to correctly identify complex finger semantics (top), handle rapid motion dynamics (middle), and disambiguate fine-grained object interactions (bottom).
    }
    \label{fig:qualitative_cases}
\end{figure*}

\clearpage
\onecolumn    % 强制切单栏
\twocolumn    % 立即切回双栏（这会起到 Reset 的作用）
{
    \small
    \bibliographystyle{ieeenat_fullname}
    \bibliography{main}
}

% X_suppl ========================

% \clearpage
% \onecolumn    % 强制切单栏
% \twocolumn    % 立即切回双栏（这会起到 Reset 的作用）
% {
%     \small
%     \bibliographystyle{ieeenat_fullname}
%     \bibliography{main}

% }
\end{document}